**Automatic Adaptation to Concept Complexity and Subjective**

**Natural Concepts: A Cognitive Model based on Chunking**

Dmitry Bennett and Fernand Gobet

London School of Economics and Political Science

Author's Note

Corresponding author: Dmitry Bennett. Email: D.Bennett5@lse.ac.uk

# Abstract

A key issue in cognitive science concerns the fundamental psychological processes that underlie the formation and retrieval of multiple types of concepts in short-term and long-term memory (STM and LTM, respectively). We propose that chunking mechanisms play an essential role and show how the CogAct computational model grounds concept learning in fundamental cognitive processes and structures (such as chunking, attention, STM and LTM). This is done in two ways. First are the in-principle demonstrations, with CogAct automatically adapting to learn a range of categories – from simple logical functions, to artificial categories, to natural raw (as opposed to natural pre-processed) concepts in the dissimilar domains of literature, chess and music. This kind of adaptive learning is difficult for most other psychological models, e.g., with cognitive models stopping at modelling artificial categories and (non-GPT) models based on deep learning requiring task-specific changes to the architecture. Secondly, we offer novel ways of designing human benchmarks for concept learning experiments and simulations – accounting for subjectivity, ways to control for individual human experiences, all while keeping to real-life complex categories. We ground CogAct in simulations of subjective conceptual spaces of individual human participants, capturing humans' subjective judgements in music, with the models learning from raw music score data without bootstrapping to pre-built knowledge structures. The CogAct simulations are compared to those obtained by a deep-learning model. These findings integrate concept learning and adaptation to complexity into the broader theories of cognitive psychology. Our approach may also be used in psychological applications that move away from modelling the average participant and towards capturing subjective concept space.

**Automatic Adaptation to Concept Complexity and Subjective Natural**

**Concepts: A Cognitive Model based on Chunking**

Research into categorisation is concerned with the fundamental question of how we ascribe objects to a particular category. Each novel experience – be it reading a new book, learning to play chess or studying a novel piece of piano music – is somehow tied to our knowledge of the world. Are the cognitive mechanisms for learning simple concepts like castling the same as for learning something more complex, like the Sicilian defence? Do we form, store, update and apply chess concepts in the same way as concepts for Bach's music or Homer's poetry? Moreover, one of the key distinguishing features of an intelligent organism is subjectivity. The same novel piece of music may be perceived as a Mozart or a Beethoven by two different people, while they both may agree that another novel piece is by Schubert. How does this come to be, what are the mechanisms that underlie this fundamental ability of ours, the ability to form concepts and categorise?

This article is concerned with an investigation of the fundamental learning mechanisms implicated in concept formation.[1] Recent trends in studying concepts shifted the focus away from scrutinizing defining features and sufficiency criteria towards probabilistic family resemblances and typicality measures. Moreover, the ongoing revolution in artificial neural networks (ANNs; a.k.a. connectionism and deep learning) sheds light on how learning mechanisms are grounded in fundamental neural structures and processes (e.g., neurons, synapses, and spreading of neural activation). We advance a model that embodies a different level of explicit representation – the cognitive, or the "mind" level – thus grounding the modern approach to concept formation in fundamental cognitive-psychology structures and processes (e.g., short-term memory, long-term memory, chunking, and attention).

_______________________

[1] In this article, we will treat "concept formation" and "concept learning" as synonymous.

The notion of grounding motivates the two main goals of the current article.  The first goal is to present a computational cognitive modelling approach that is – in important ways – analogous to deep learning. Despite having a different set of assumptions and proposed mechanisms, our cognitive model can also account for categorisation in realistic, highly complex domains, all while learning from raw data without bootstrapping to experimenter-devised feature detectors and knowledge structures. We will discuss the functional similarity between cognitive chunks and ANNs' neural engrams. Another key feature is that our model automatically *adapts* to task complexity and does not need ad hoc changes to the architecture to progress from rote learning, to simple artificial concepts, to highly complex real-life natural concepts in the dissimilar domains of literature, music and chess.

The second goal is to develop a novel framework for human benchmarks for computer models of concept formation. Our implementation also further grounds our model with human data. Traditionally, human performance data tended to be averaged over many trials and participants, bypassing the issue of individual differences and *subjective* judgment, as well as affording more scope for the models to be correct. Moreover, attempts to control participants' "training data" have often relied on artificial concepts (such as novel geometrical shapes). A key contribution of the current work is that we measure and model the subjective conceptual apparatus of individual participants, as opposed to the traditional "average subject" approach. We also develop a method which allows us to control for human participants' "training data," while keeping to complex real-life stimuli. Finally, we present a novel way of measuring the fit of formal psychological models to human data, with five progressively more stringent metrics that go beyond the traditional statistical-significance calculations.

## Theoretical Background

### Bespoke models of categorisation

The work in this article draws from two traditions in psychological research: bespoke theories/models of categorisation and broader cognitive architectures known under the umbrella term of unified theories of cognition.

Initial theories of categorisation focused on rules and definitions (Bruner, Goodnow, & Austin, 1986; Hull, 1920; Smoke, 1932), but seminal work by Wittgenstein (1953), Posner and Keele (1968), Rosch (1973), and Medin and Schaffer (1978) showed that natural categories lacked consistent defining features and were better characterized in terms of probabilistic structures. Two ideas – Rosch's prototypes and Medin's exemplars – turned out to be particularly dominant; this is evident from the chapters on concept formation of any standard cognitive psychology textbook (Eysenck & Keane, 2005; Gobet, Chassy, & Bilalic, 2011; Goldstein, 2014; Harley, 2014; Margolis & Laurence, 2015; Murphy, 2013; Reisberg, 2019).

Prototype theory suggested that each concept is represented by an abstract prototype – a set of *characteristic* features. These features are probabilistic, with some being absent altogether while others being differentiated by their weighting/importance. Category membership is then calculated by evaluating stimulus' similarity to the prototype (Bowman, Iwashita, & Zeithamova, 2020; Lupyan, 2017; Rosch, 1975).

The exemplar theory was motivated by the individual instances of category members (also called exemplars) that a person encounters during her lifetime experience (Medin & Schaffer, 1978; Shin & Nosofsky, 1992; E. E. Smith & Medin, 1981). It posits that the memory system stores large numbers of specific instances and derives a typicality gradient from their overall population. This implies that, for each test, prototype theory derives a single concept space distance (i.e., from stimulus to the prototype), but exemplar theory derives many such distances (i.e., from stimulus to each of the exemplars).

In the decades following their emergence, these two theories have generated immense amounts of research, with the exemplar theory having the upper hand – particularly in studies where artificial categories had very complex boundaries and were thus difficult to learn (McKinley & Nosofsky, 1995; Murphy, 2016; D. J. Smith & Minda, 2000; Storms, Boeck, & Ruts, 2000). For example, in a study with five participants learning two artificial categories with highly non-linear classification boundaries (circles with varying diameter and radial line orientation), it was shown that people can learn them with sufficient training and that exemplar model provides better fit to the human categorisation performance when compared to alternative models (McKinley & Nosofsky, 1995). In another example, a psychological model of categorisation of geological rocks was proposed, demonstrating that the exemplar theory was able to scale up to natural categories (Nosofsky, Meagher, & Kumar, 2022; Nosofsky et al., 2017). Classification of natural rock specimens was made possible by participants providing similarity judgments among the pairs of rock stimuli which were then fed into an 8-dimensional multidimensional scaling (MDS) solver, or by deriving corresponding dimensions from geology textbooks. The latter process resulted in a mixture of 17 psychological dimensions: some being continuous (e.g., participants' ratings ranging from 0 to 9, describing such rock characteristics as "lightness of colour", "shininess", "roughness") and others being binary (i.e., participants' ratings of some features of rocks as being either present in or absent from the presented stimuli – such as "stripes" or "holes"). In a later extension of this study, a deep learning model was trained to extract the MDS coordinates of rock images (once again, by using human pair similarity judgments to train the model) (Sanders & Nosofsky, 2020).

A family of approaches known under the umbrella term "clustering models" extend the abstraction approach of the exemplar and prototype theories, but with other types of clustering – e.g., mechanisms based on Bayesian inference (Anderson, 1991, 2010; Konovalova & Le Mens, 2018) and adaptive learning (Love & Medin, 1998; Love, Medin, & Gureckis, 2004; Mok & Love, 2019; Vanpaemel & Storms, 2008). For example, the adaptive learning SUSTAIN (Supervised and Unsupervised STratified Adaptive Incremental Network) models its representations as intermediate between the prototype-exemplar extremes, with each category being generally represented by a few clusters of objects rather than all exemplars or a single prototype (Love & Medin, 1998; Love et al., 2004).

We will highlight two important principles of SUSTAIN. First, it developed an adaptive approach to matching the complexity of the learner to that of the learning problem, a psychological solution to a machine learning problem known as the bias-variance dilemma (Geman, Bienenstock, & Doursat, 1992; Love et al., 2004). In brief, when a simple model is matched with a complex task boundary, it ends up underfitting and cannot learn it due to possessing too much bias. When a complex model is tasked to learn a simple boundary, it ends up overfitting and cannot learn it due to possessing too much variance. The dilemma may also be summed up in the form of a simple example: a 10 neuron ANN does well with AND, OR, XOR binary logic problems, but cannot work with real-life images. The opposite is true for a 1000 neuron ANN – it can categorise real-life images like hand-written digits, but it loses the ability to learn simple binary logic. Common "solutions" to the dilemma include setting the model complexity in accordance with the task (e.g., choosing the architecture and the number of hidden units of an ANN model) and/or increasing the training data size towards infinity (Geman et al., 1992). SUSTAIN solves this problem by incrementally growing cluster representations during learning. New clusters are developed only in response to a surprising event (e.g., a prediction error, or encountering an object that is too dissimilar to any existing clusters). Categorisation involves activation of the most dominant of these clusters.

Another important contribution of SUSTAIN is that it is economical in its storage and creates a new separate cluster only following a prediction error. The economy principle is also carried into categorisation: SUSTAIN makes predictions by focusing on one winning cluster (Love & Medin, 1998; Love et al., 2004), as opposed to all of the stored exemplars (Nosofsky, 2011; Shin & Nosofsky, 1992), or the Bayesian approach that computes weighted sum over all of the stored clusters (Anderson, 1991). The narrow focus on the winning clusters may be a more realistic simulation of human performance (Murphy & Ross, 1994, 2010). (To pre-empt the sections that follow, our CogAct categorisation model also adheres to the adaptation and economy principles championed by SUSTAIN, albeit based on different mechanisms and processes.)

Despite the insightfulness of the work above, it was subject to three key limitations. First, it focuses on low-dimensional hand-coded or low-dimensional MDS stimuli – that is, artificial or natural pre-processed stimuli. Indeed, both prototype, exemplar and clustering categorisation models competed with each other in experiments with highly simplified categories (e.g., Anderson, 1991; Braunlich & Love, 2022; Love et al., 2004; McKinley & Nosofsky, 1995; Nosofsky et al., 1994; Nosofsky, Sanders, Meagher, et al., 2018; Posner & Keele, 1968; Reed, 1972; Shepard, Hovland, & Jenkins, 1961; D. J. Smith & Minda, 2000; Vanpaemel & Storms, 2008). This allowed experimenters to create workable approximations to the inaccessible psychological feature space that contained the categorical mental representations (i.e., concepts). However, this also led to concerns about ecological validity: highly controlled and simplified stimuli may not be closely related to the complex multidimensional environment that they are meant to represent. For example, while it is an accepted conclusion that artificial categories with complex boundaries are better described by the exemplar theory and not the prototype (see above), this conclusion no longer holds true when applied to categorisation of natural images. Indeed, this was one of the major findings of a visual categorisation study that involved deep-learning algorithms trained on 10,000 natural images across ten categories and over 500,000 human categorisation decisions (Battleday, Peterson, & Griffiths, 2020). Their broader conclusion was that intuitions about theory and model performance for low-dimensional categories do not transfer to higher-dimensional ones.

Preprocessing natural categories also brings a second issue: bootstrapping models by human knowledge. Whether using human experts' pairs of similarity judgments to train the categorisation models directly (Nosofsky, Sanders, Meagher, et al., 2018), or using human generated pairs of similarity judgments to train ANNs that are then integrated with the categorisation models (Sanders & Nosofsky, 2020), both negate an important question: *how* do human rock geology experts make their pair-wise judgments? This vital part of the categorisation question is eschewed by the models' bootstrapping to human expertise which the models are then tasked to explain. A continuation of this problem is domain specificity. Even a hypothetical perfect model of categorising rocks would be completely helpless with categorising cats (cars, poetry, music, chess positions … ad infinitum) – until datasets with corresponding human pair-wise judgements are available. This has direct implications for the claim of such theory to capturing *general* concept learning.

The third limitation is that – bar a few exceptions (e.g., McKinley & Nosofsky, 1995) – categorisation models tended to fit to data that were averaged over participants. This may be misleading as such data may obscure categorisation patterns at the level of the individual participant (Gobet, 2017; Newell, 1973; Vanpaemel & Storms, 2008). (To pre-empt the sections that follow, our CogAct categorisation model will simulate participants at the level of the individual. It will also learn both low-dimensional artificial and high-dimensional natural categories without bootstrapping to pre-learned knowledge structures.)

## Unified theories of cognition and cognitive architectures

Apart from bespoke theories and models of categorisation, our article also builds off a second area of research in psychology – unified theories of cognition (UTCs). The idea of a UTC originated from criticism of the traditionally disjointed approach to cognitive science and its task-specific models (Newell, 1973). UTCs were presented as a solution to that problem, integrating theories and models of various cognitive mechanisms and representations into a single cognitive theory/architecture (Byrne, 2012; Lane & Gobet, 2012b; Newell, 1990). Thus, while a cognitive *model* usually considers a single behaviour (such as categorisation), the aim of a cognitive *architecture* is to capture a wide array of intelligent behaviours into a single set of interlinked processes (Anderson & Lebiere, 1998; Laird, Lebiere, & Rosenbloom, 2017; Newell, 1990; Ritter, Tehranchi, & Oury, 2019). The UTC idea is linked to the "Single Algorithm Hypothesis" that proposed that visual, auditory, motor, and somato-sensory brain regions utilise approximately the same algorithm to extract approximately one type of data structure from various types of information (Hawkins & Blakeslee, 2004; Mountcastle, 1978). Also, UTC is related to artificial general Intelligence (AGI) – a single algorithm that has a wide range of competencies across a varied range of tasks (Mnih et al., 2015; Russell & Norvig, 2021; Silver et al., 2017; Swan et al., 2022).

Many different cognitive architectures have been proposed, with about 50 currently active (Kotseruba & Tsotsos, 2020). While this number is large, Newell argued that this is where psychology should concentrate its efforts, instead of continuing to add to the (as of then) around 10,000 established empirical regularities which have not yet been integrated into a general theoretical structure (Newell, 1990, p. 237). The wide range of competing architectures arose from this type of reasoning, each focused on different kinds of behaviour and levels of analysis. For example, Soar (Newell, 1990), EPAM (Feigenbaum & Simon, 1984) and CHREST (Gobet, 1998) focus on symbolic information processing; ART (Grossberg, 1999) is an example of connectionist sub-symbolic approach; while ACT-R (Anderson & Lebiere, 1998) uses a hybrid approach that fuses the two approaches above.

With each UTC predicting and explaining multiple psychological phenomena with a single set of overarching processes, concept learning/categorisation mechanisms were also proposed. For example, a categorisation model was integrated in Soar architecture and was able to convert linguistic input like "feline with mane" into a category output like "lion" (Lieto et al., 2017). The advantage of this model is that apart from its outright categorisation performance, it was constrained by Soar's task-independent mechanisms like working memory, semantic memory, procedural memory and chunks (Laird, 2022).

A different concept learning model – focused on categorising chess openings[2] – was built using a CHREST cognitive architecture (Lane & Gobet, 2012c). The model was once again largely based on the task-independent cognitive mechanisms – this time developed by CHREST/template theory (Gobet & Simon, 1996c, 2000), which in turn extended the EPAM/chunking theory (Chase & Simon, 1973). (To pre-empt the sections that follow, our CogAct model will also follow the chunking theory tradition; we will discuss it in detail in subsequent sections of the article.)

Apart from being integrated into broader psychological theories, one important strength of these two categorisation models is their move away from low-dimensional artificial stimuli and towards real life complexity of natural categories.

On the other hand, they also share a common problem: the reliance on hand-crafted features. For instance, in the case of the Soar categorisation model, Babel Net and Concept Net dictionaries were used (Lieto et al., 2017). Thus, the "dog" entry has database links like *HasParts* legs, *CapableOf* guarding property, *UsedFor* companionship, *RelatedTo* wolf, *NotDesires* fleas. In the case of the CHREST's chess categorisation model, chess-specific hand-crafted heuristics were devised. For example, one of the heuristics guides model's attention towards chess pieces under attack.

―――――――――――――

[2] A chess opening is a typical pattern for approximately the first ten to twenty moves of a chess game. It is typical for expert players to study opening in depth, despite there being billions of possible sequences to begin a game.

This leads to three fundamental problems. The first is that the within-domain performance of the model depends on the extensiveness of the dataset and may never capture even a single natural category with its (near infinite) variability. For example, if the *CanSleepUnder* relationship was not coded into the "dog" concept training dataset, the model would not be able to categorise sentences about animals sleeping under radiators as members of the dog category.

The second problem – domain specificity – was discussed in the context of exemplar and clustering models, but also applies here. Indeed, even a hypothetical perfect model of categorising dogs or chess positions would be completely helpless with categorising cats (cars, poetry, music … ad infinitum) – until datasets with corresponding handcrafted features were available. This has direct implications for the claim of such theories to capturing *general* concept learning.

The third question of who does the heavy lifting in explaining the results of categorisation experiments – the model or the human dictionary writers? – is once again apposite. *How* do human experts make dictionary entries and form chess heuristics seems to be a vital part of the question that is negated by the models' bootstrapping to prebuilt knowledge structures.

## A Cognitive Model of Categorisation Based on Chunking

We will now present our model of categorisation – CogAct (Chunking Organisation Grouping Activator, or, more simply, Cognitive Activator). As was mentioned above, our model extends a UTC/cognitive architecture. It also adheres to the adaptation to complexity and economy principles, in a similar way to the SUSTAIN model (Love et al., 2004) discussed above. However, our model does not stop at artificial low-dimensional stimuli, but applies the same principles to learn high-dimensional natural categories without bootstrapping to pre-learned knowledge structures. We apply CogAct to a range of categorisation problems, starting from rote learning, moving to artificial concepts, and then to dissimilar natural concepts in literature, music and chess. This addresses the recent criticism of categorisation models projecting findings with low-dimensional stimuli to high-dimensional natural categories (Battleday et al., 2020). We conclude by grounding our model in simulations of human subjective natural concept spaces simulated at the individual level and comparing it against a deep learning baseline. The mechanisms will be the same in all cases, as the model will adapt itself to domain and task complexity.

The current section provides a brief introduction to CogAct and chunking theory – a UTC which underlies it. We then describe CogAct structures and mechanisms, including simple symbolic pattern manipulation.

## Chunking theory and CogAct

A *chunk* can be defined as a meaningful unit of information constructed from elements that have strong associations between each other (e.g., several digits making up a telephone number or a group of letters and digits making up a postal address), and *chunking* is the process of creating and updating chunks in the cognitive system (Gobet, Lloyd-Kelly, & Lane, 2016; Simon, 1974). Although the chunks themselves vary between people due to personal differences, chunking mechanisms are largely invariant across domains, individuals, and cultures (Chase & Simon, 1973; Gobet et al., 2001; Miller, 1956; Simon, 1974).

Since their emergence in 1959, the models based on chunking were used to predict and explain behaviour in verbal learning research (Feigenbaum, 1959; Feigenbaum & Simon, 1984; Richman & Simon, 1989; Richman, Simon, & Feigenbaum, 2002), accounted for context effects in perception, various aspects of concept formation (Lane & Gobet, 2012a, 2012c), problem solving (Lane, Cheng, & Gobet, 2000), language acquisition (Freudenthal et al., 2016), emotion processing in problem gambling (Schiller & Gobet, 2014), developmental abilities and cognitive decline due to ageing (Mathy et al., 2016; R. Smith, Gobet, & Lane, 2007), and shed light on the memory systems involved in expert behaviour (Gobet & Simon, 2000; Richman et al., 1996; Richman, Staszewski, & Simon, 1995; Simon & Chase, 1973; Simon & Gilmartin, 1973).

All of the research above (plus more) was encapsulated into a continuously updated lineage of computational cognitive architectures: first as EPAM (Elementary Perceiver And Memorizer) (Feigenbaum, 1963; Richman et al., 1995), then CHREST (Chunking Hierarchy REtrieval STructures) (Gobet, 1993, 2000; Gobet & Lane, 2012; Gobet & Simon, 2000), and MOSAIC (Model of Syntax Acquisition in Children) (Freudenthal et al., 2007; Freudenthal et al., 2016). CogAct is to continue this lineage.

The simple stimuli used in verbal learning research provide for a good introduction to chunking theory. For example, Richman et al. (2002) reported that these models have captured at least 20 regularities identified by research into human rote learning. In this line of research, participants are taught lists of paired stimulus-response (S-R) made of nonsense consonant trigrams in an effort to uncover the key laws of learning. EPAM successfully simulated the "intralist similarity" effect: humans (and EPAM) produced more errors in S-R learning trials when the nonsense trigrams were similar to each other (e.g. [Z,X,K] and [Z,H,J]) (Hintzman, 1968; Richman et al., 2002). It also simulated the primacy-recency serial position curve that describes people's tendency to remember the first and last items of a list better than the items in the middle (Feigenbaum & Simon, 1962; Richman et al., 2002). In the next sections, we will also use the nonsense trigram verbal learning stimuli to explain the foundations of the current CogAct model. We will demonstrate how the model automatically upscales to process real-life natural domains. Thus, mechanisms underpinning CogAct are not arbitrary but are informed by verbal-learning results, which CogAct can model too.

**The Present CogAct Model**

CogAct is a self-organising computer model that simulates human learning processes. Broadly speaking, CogAct consists of four interconnected modules: the environmental input module, STM, LTM, and model's output/behaviour module (see Figure 1).

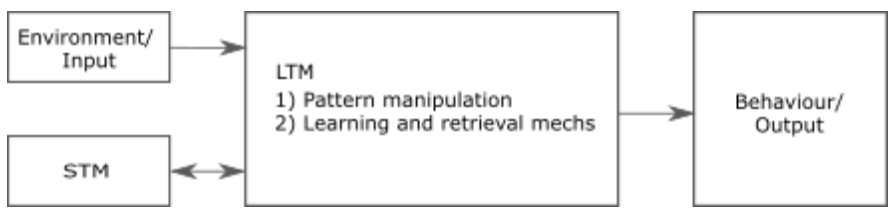

*Figure 1. A bird's-eye view of CogAct.*

***Environmental input module***

The environmental input module is where the sensory organs (e.g., the eye or the ear) of the organism operate, although no detailed mechanisms are implemented in the current version of the model. The patterns that are processed by CogAct are

symbolic – they are meaningful and are represented in identical ways for objects inside (cognition) and outside (input) the architecture. Patterns are assumed to be composite objects made up of primitives: for example, a collection of letters making up words, a collection of words making up sentences, a collection of chords making up musical measures, or a collection of measures making up a musical piece. While similar to deep learning in its focus on perception and learning as the primary attributes of cognition, CogAct is an example of a symbolic architecture and thus contrasts with the connectionist neural network approach, which is sub-symbolic. The focus on symbols is natural, given CogAct's roots in cognitive psychology.

### Long-Term Memory

LTM is the most complex part of CogAct. It is also central to all of its behaviours – debilitating LTM progressively affects the way CogAct responds to stimuli, just like with neural networks. The LTM may be split into two interacting parts. The first part is concerned with manipulation of patterns and is essential to all CogAct processes. It has three inbuilt functions: one that tests if two patterns are *equal*, another that tests if they *match* and the third function finds the *difference* between two patterns.

The patterns are found to be *equal* if they are comprised of exactly the same primitives. For example, [A,B] is *equal* to [A,B]. The patterns are said to *match* if their sequences overlap. For example, [A,B,C] *matches* [A,B,C,D] but does not *match* [A,C,B]. Lastly, the *difference* between two non-matching patterns is computed as well. It returns the part of the first pattern after the common part of the compared patterns is removed. For example, the *difference* between [A,B,F,C] and [A,B,C] is [F,C]; the *difference* between [A,B] and [B,C] is [A,B].

The rest of this section is concerned with the (much more complex) second part of the LTM which brings about learning and retrieval of knowledge. One of its key features is a *discrimination network*, which consists of *nodes* that are connected by *test links*. The main purpose of the discrimination network is to sort an input pattern to the most relevant branch of the LTM (it also has secondary roles, like forming associations between different parts of the LTM, and these will be explicated further below). Thus, the discrimination network serves both a learning/retrieval device and as a mechanism akin to a similarity function – somewhat analogous to the hidden layers of a deep-learning ANN (Lane & Gobet, 2012a).

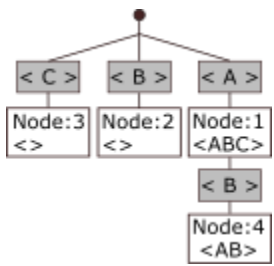

*Figure 2. An example of CogAct's LTM discrimination network. Each of its four nodes contains a test link (e.g., "B") and an image (e.g., "AB").*

In this network, the test links are comprised of patterns that are checked against the current input during retrieval; the nodes contain *images* with *familiarised* parts of patterns. Images represent the information that CogAct can output when a node is accessed. It should be noted that the images may be smaller than, equal to, or bigger than the tests required to reach each node. For example, Node 1 in Figure 2 has test link ['A'] and the larger image ['A', 'B', 'C'].

Thus, the definition of a "chunk" can now be formalised: the series of tests required to reach a particular node (also known as node's *path* or *contents*) form the *extrinsic description* of a chunk, while the node's image forms the *intrinsic description* of a chunk.

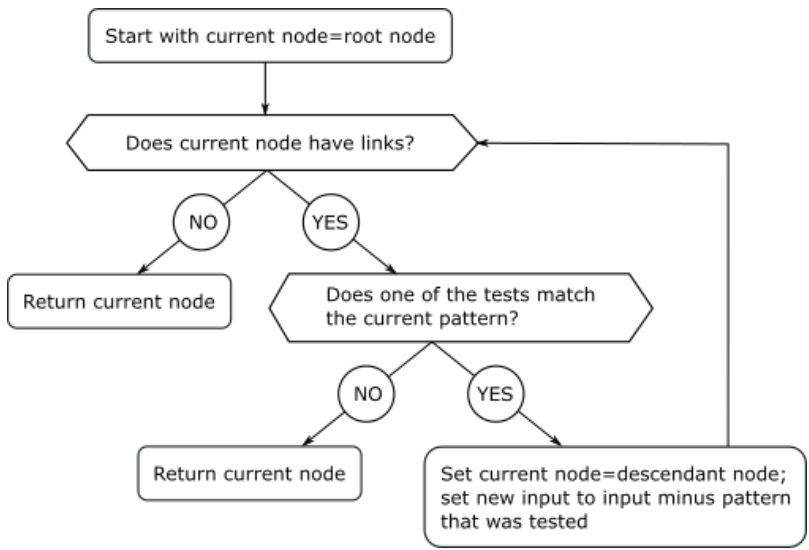

*Figure 3. The flow chart representing the recognition / retrieval of a chunk from CogAct's LTM.*

Retrieval of a chunk from the network is performed by CogAct's *recognise* function, which is described by the flow chart in Figure 3. This function takes in a pattern and outputs (*retrieves*) a node if some form of recognition is possible, or the root node, if that is not the case. For example, if the pattern [D] was presented as input of a model with the LTM tree of Figure 2, CogAct would retrieve [root], i.e., it would not recognise it as anything. On the other hand, input patterns [A,B] and [A,B,C] would be recognised as [A,B].

To sum up, chunk retrieval process can be summarised as follows. Upon receiving a pattern, CogAct begins sorting it through the network, starting with the current node (set to root node in the beginning). If the current node has test links and if one of its test links matches the input pattern (via the aforementioned *match* function), this node becomes set to current node, the new test pattern becomes the original tested pattern minus the previous test pattern, and the loop continues until the descendant node either does not have test links or neither of its test links match the input pattern. At this stage, the node is retrieved, and its chunk becomes available for further analysis and/or learning process.

*The learning processes.*

CogAct simulates human learning by expanding its LTM network, where each node is comprised of a branch/test link and a leaf/image. While the *test links* check for the presence of perceptual features, the *images* form internal representations of the external stimuli. Learning takes place by comparing the information held in the images with that of the stimulus. If the two match perfectly, no learning happens. If there is a mismatch and the image is a subset of the stimulus, further information is added to the image. In all other cases, additional test links and nodes are added either to the root node or to the internal nodes, with the possibility of adding new information to the nodes already present. Concretely, chunking theory posits that the learning process is spread over four mutually integrated stages (Gobet & Lane, 2012):

1. Pattern is sorted via the retrieve mechanism to a node.

2. The image of the reached node is compared with the external pattern.

3. If the two match (according to the *match* function), *familiarisation* takes place.

4. If the pattern and the retrieved image do not match, *discrimination* follows.

In other words, during learning, *familiarisation* increases the amount of data that CogAct can retrieve from a particular chunk, while *discrimination* increases the number of chunks that CogAct can identify (Gobet et al., 2001; Lane & Gobet, 2012a).

*The discrimination process.*

Discrimination is the process of adding new nodes to the LTM network, parametrised by the current node and an input pattern (as passed by the *learn* function).

For example, if our LTM contains a single node with a test link ['A'] and an image ['A'], learning a new primitive ['B'] would imply adding a new node with the corresponding test link to the network (see Figure 4 and also Figure 5 for an example with a bigger initial state LTM tree).

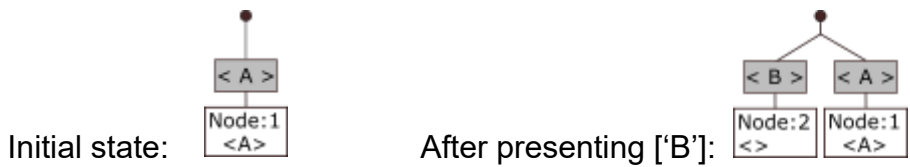

Initial state:          After presenting ['B']:

*Figure 4. Discrimination process in action, represented pictorially. Note that while the added 2nd node has a test link ['B'], its image is empty.*

Initial state:

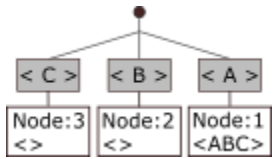

After presenting ['A','B']:

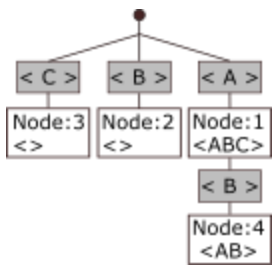

*Figure 5. Another example of the Discrimination process. Note that in this case one single "step" of discrimination has created a node with both a test link ['B'] and also a filled image ['A', 'B'], since the primitives A and B are already known. The latter is an example of how CogAct represents "single-shot learning".*

More concretely, the *discrimination* function is parametrised by the chunk node that is passed in from the *learn* function and the input pattern. During the *discrimination* process, CogAct takes the difference part of the input pattern (generated by the *difference* function) and attempts to sort it through the existing network via the *recognise* function. If this attempt leads to staying at the root of the memory tree, LTM learns/develops a new primitive. Conversely, if the retrieved node is not the root, it considers the image of the retrieved node. If the image is empty, CogAct integrates/*familiarises* the difference pattern. If the image is not empty, CogAct creates a new node. Its test link would incorporate the retrieved (via the *recognise* function) image without the end marker, while its image would become the set of tests needed to reach that node.

> *The familiarisation process.*

If discrimination concerns adding a new node to the LTM network, *familiarisation* is about adding new information to an existing chunk. If the perceived pattern contains not just everything that is already present in the retrieved chunk, but some extra on top, then familiarisation adds extra information to the chunk. As noted above, familiarisation increases the amount of data that can be retrieved from a chunk, while discrimination increases the range of chunks that can be identified (Gobet & Lane, 2012; Gobet et al., 2001).

More concretely, the *familiarisation* function takes the *difference* between the pattern and the retrieved image and sorts that difference through the network (via the *recognise* function), generating four potential outcomes.

The first option is finding no difference between the retrieved image and the input pattern; in this case familiarisation does nothing at all. The second possibility is retrieving the root node, this triggers discrimination and subsequent learning of a new primitive. The third prospect is finding the retrieved image to be empty or being larger than the difference; in this case the first primitive from the found difference is added to the image of the original node (i.e., the node that was passed into familiarisation function by the learn method). Lastly, in all other cases, the image that is retrieved by sorting the difference pattern is augmented by the first item of that difference.

For example, if the LTM network contains a node with a test link ['A'] and an empty image, subsequent presentation of any pattern that begins with ['A'] would trigger familiarisation and add ['A'] to the lone node's image (see Figure 6 for the simpler case with an empty initial image and Figure 7 for an example of image / chunk extension).

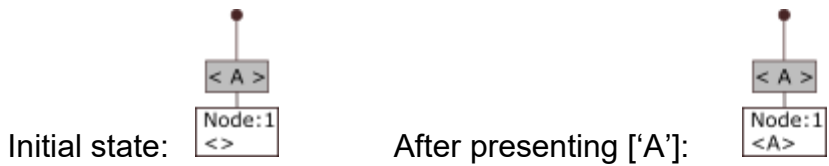

Initial state:                    After presenting ['A']:

*Figure 6. Familiarisation process in action, represented pictorially. An empty image of the 1st node (containing test link ['A']) is filled with an image ['A'].*

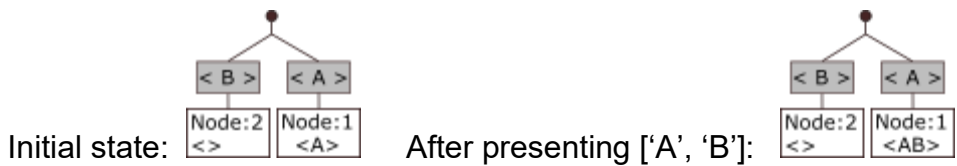

Initial state:                    After presenting ['A', 'B']:

*Figure 7. Familiarisation process in action, represented pictorially. An image of the 1st node ['A'] is appended with a primitive / feature ['B'] to form the ['A', 'B'] chunk.*

The last important point about the LTM is that it may have multiple discrimination networks at the same time, one network per modality. While modalities have different names (e.g., Visual and Verbal, see *Figure 8*), the chunking mechanisms are identical for all of them.

### *Short-Term Memory*

Having outlined the key mechanisms behind CogAct's operation of the LTM, we can now switch to the part of the architecture that concerns modelling short-term memory (STM). As we shall see, STM forms a crucial link in the way it integrates with the input-output mechanisms and the LTM itself. Further, STM is also fundamental in the way it enables other, not yet discussed, powerful mechanisms of the LTM – with examples including similarity links, naming links and templates to mention but three.

Broadly speaking, STM is a capacity-limited and temporary storage space, with more recent input displacing earlier input (Gobet et al., 2001). STM holds *pointers* to LTM's nodes (operationalised as linked lists data structures – see below). Lastly, different modalities in CogAct have their own STMs (for example, verbal, visual, and action types of patterns that are stored and processed by separate respective LTM networks also use separate STMs).

Concretely, CogAct's STM is a queue-like (first in, first out) linked list data structure. It has finite length (determined by a hyper-parameter to account for learners' individual differences and typically set between 2 and 9 chunks per modality).

The main body of the discrimination network is also connected to STM mechanisms. Specifically, the *learn* and *recognise* functions feed the LTM nodes into the STM at the end of their cycles.

*Lateral naming links*

As was mentioned above, STM also enables a range of powerful new tools for the LTM architecture. The naming links are particularly crucial to the concept formation tasks. Lateral link learning can only happen when CogAct's STM contains pointers to some relevant nodes of the LTM (see Figure 8). This implies that nodes are linked based on temporal or spatial contiguity (e.g., chunks of musical melodies as an example of the former, chunks of chess positions as an example of the latter, sequential chord structure or a sequence of chess moves being examples of both) and that this contiguity is being preserved. The use of lateral links underlies supervised learning in CogAct (i.e., assigning labels to data), while unsupervised learning is done via other chunking mechanisms outlined above. CogAct automatically switches between these two modes of learning depending on the presence of simultaneous modality overlap in its input stream (e.g., Visual and Verbal, or Visual and Action and so on). We should also note that, if STM or the lateral link creation functions were disconnected/debilitated, recognition would continue to function, but further categorical learning would become impossible for CogAct – similar to the famous clinical case of patient H.M. (Squire, 2009) .

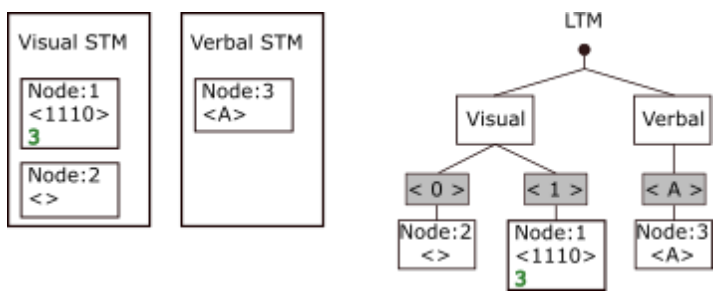

*Figure 8. Simultaneous presence of fully learned chunks (Node 1 and Node 3) in different STM modalities triggers creation of a lateral naming link between the chunks.*

***Attentional and chunk activation mechanisms***

Attention and selective LTM activation form an important part of CogAct. Firstly, a recursive "sliding attention window" represents the limited scope of a human reader and allows multiple passes over patterns deemed unfamiliar by the LTM. If a vector of patterns can be represented as input ***p*** = $[p_1, p_2, \dots p_n]$, then CogAct attention window ***w*** would fetch $[p_{1+t}, p_{2+t}, \dots p_{m+t}]$, with $m$ being the span of the window and $t$ being the time step size. Before proceeding to the next step, the window would progressively shrink and fetch sequences $[p_{2+t}, \dots p_{m+t}]$, $[p_{3+t}, \dots p_{m+t}]$, $\dots [p_{m+t-1}, p_{m+t}]$.

Secondly, CogAct records the "activation" of a chunk – the largest chunk met so far – as a function of an input pattern, to allow for conflict resolution between chunks "voting" for different categories.

If there are $m$ categories, the vector of category labels is ***c*** = $[c_1, c_2, \dots c_m]$, the vector of category specific chunk activations is ***a*** = $[a_1, a_2, \dots a_m]$ and the confidence score of a prediction that a pattern belongs to category $c_i$ would be calculated using the equation

$$C(c_i|x) = a_i \ / \sum_{k=1}^{m} (a_k)$$

where $C(c_i|x)$ is confidence that category label is $c_i$, given a book or music score $x$; $a_i$ is the LTM chunks' activation corresponding to that category, and the summation part is the sum of chunk activations across all $m$ categories. For example, having been given a novel Beethoven music score, the model may be 60% confident that is a Mozart, 30% confident that it is Beethoven and 10% confident that it is Bach.

This mechanism was inspired by the review of neuroimaging research by Guida et al. (2012a), who found experts to possess large domain-specific knowledge structures that activate in the areas of the brain associated with episodic LTM memory. While novices primarily rely on the prefrontal cortex to form new primitives and update their shallow chunking networks, experts show less activation in the prefrontal areas, but large activations in the medial temporal lobe, presumably due to rapid utilisation of large knowledge structures (Guida, Gobet, & Nicolas, 2013; Guida et al., 2012a).

The retrieved verbal and visual LTM chunks send pointers to STM, which then enables naming links to be stored in LTM. It is important to stress that the length of the STM queue is measured in chunks and not in primitives – as we can, for example, briefly memorise 7(+/-2) individual digits, but also 7(+/-2) previously learnt telephone numbers (Miller, 1956).

### *Behaviour/Output module*

CogAct responds to specific commands like *categorise*, *learn*, and *retrieve*. The *learn* command triggers the learning process involves the growth of the LTM via the aforementioned *discrimination* and *familiarisation* processes by processing the data passed into the Environmental input module. This data can be either labelled or unlabelled. The *categorise* command triggers CogAct to search for a label to a stimulus provided in the Environmental input module. The *retrieve* command causes CogAct to compare the input pattern of the Environmental input module to its LTM contents and produce a chunk that is the most similar to the input.

## CogAct's Adaptation to the Variety and Complexity of Categories

In the current section, we will show how CogAct automatically scales its mechanisms to learn gradually more complex categorisation problems. We will begin with learning a simple logical exclusive OR function (XOR) and proceed to artificial concepts (with few binary dimensions) and conclude with full complexity of real-life categories. In all cases, CogAct's architecture will remain unchanged from problem to problem. The main idea behind this section is to resolve the in-principle objections to what could otherwise seem like domain- and task-specific mechanisms of CogAct that simulate human behaviour in the later section of the article. Also, having taken the criticism of low-dimensional research (Battleday et al., 2020) on board, we nonetheless wanted to form a bridge to the vast literature on low-dimensional concept learning, as well as present a way of how intuitions about theory and model performance on low-dimensional stimuli *can* be transferred to high-dimensional natural categories.

### Learning the XOR function

Learning simple logical functions (like AND, OR, XOR and so on) occupies a special place in history of formal psychological models. Neural network modelling pioneers McCulloch and Pitts (1943) demonstrated how their idealised neurons can perform basic logic functions like *AND*, *OR*, *NOT* and suggested that the brain would then produce all of its cognitive and behavioural complexity by combining these simple neural calculations, just like simple logical propositions were combined to produce complex mathematical theorems and proofs in *Principia Mathematica* (Whitehead & Russell, 1911). Computer scientists Minsky and Papert demonstrated that a perceptron model was not capable of learning the XOR function and convinced many researchers to abandon the ANNs field for over a decade (Haykin, 1999; Minsky & Papert, 1969).

More recently, and as discussed above, it was shown that, while simple models may learn simple class boundaries and complex models may learn complex class boundaries, it is difficult to criss-cross model and boundary complexities. For example, while a deep learning model made up of 10 neurons can easily learn the logical XOR function, this is difficult for a model with 1000 neurons (Geman et al., 1992; Love & Medin, 1998; Love et al., 2004).[3]

In this context, we present how a CogAct model learns XOR. The stimuli for this experiment are two-dimensional and binary valued (e.g., [1, 0]). The category label has a single dimension and is also binary valued, albeit coded in a letter form (e.g., [T] denoting True).

With such a simple task, CogAct automatically employs rote learning to form the symbolic representation of XOR. It learns the binary patterns such as [1,0] in the Visual node of LTM, learns the verbal True/False labels [T], [F] in the Verbal node of the LTM and assigns lateral links between the nodes when they coincide in the STM during the learning process (see Figure 9). Thereafter, presentation of an unlabelled visual stimulus triggers a verbal label response. For example, [1,1] returns [F], while [1, 0] returns [T]. We should note that this experimental structure will hold true for all the experiments that follow, with the difference that the number of categories, dimensionality and the values of the stimuli will progressively increase/become more complex.

---

[3] This difficulty may have been addressed in the most recent large language Generative Pretrained Transformer (GPT) ANN models. For example, the same GPT4 model is able to answer questions on simple arithmetic as well as complex moral philosophy issues without ad hoc changes to its architecture. The extremely large training data size may have thus "solved" the bias-variance dilemma – indeed, this was predicted by Geman et al. (1992). With that said, GPT4 is also susceptible to blunders in simple problems (like basic arithmetic – when not relying on a separate calculator module) (Bubeck et al., 2023).

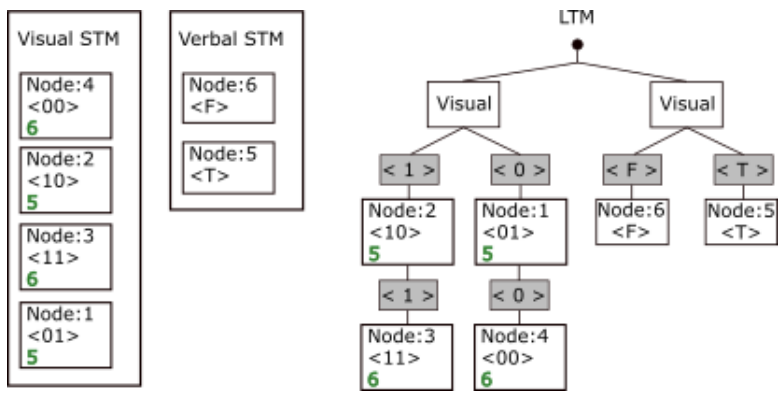

*Figure 9. Learning the XOR function. The numbers in green denote the naming links that connect the visual and the verbal nodes.*

## Artificial binary category learning

Artificial categories play an important part in the field of formal concept formation models. In some cases, the artificial categories were created from scratch by the experimenters (e.g., distorted geometrical shapes that varied along several dimensions (Anderson, 1991, 2010; Konovalova & Le Mens, 2018). In other cases, natural categories are pre-processed[4] so as to vary along several dimensions (e.g., photographs coded into four binary dimensions: hair colour, shirt colour, smile type, and hair length; and geological rock samples coded along dimensions such as grain size, opacity and so on (Love et al., 2004; Nosofsky, Sanders, & McDaniel, 2018)).

The canonical "five-four" artificial category structure forms another such example, one that is heavily featured in psychological models of concept learning to this day (Braunlich & Love, 2022; Schlegelmilch, Wills, & von Helversen, 2022; D. J. Smith & Minda, 2000; E. E. Smith & Medin, 1981). We will also use it to demonstrate that our model is similarly capable of such learning. The experiment was previously done with another chunking theory cognitive architecture – CHREST (Lane & Gobet, 2013). However, their simulation was task specific and could not adapt to other categorisation tasks. We made no adjustments to CogAct mechanisms, which remain exactly the same as in the XOR study above.

---

[4] This issue is frequently glossed over in standard cognitive psychology texts that talk about – for instance – exemplar models deriving similarity gradients of birds from the exemplars of common sparrows and rare penguins. Despite their "exemplar" label, the models never deal with the literal exemplars of rock specimens or their images. Instead, those models' process human derived scores/dimension data that *describe* the exemplars.

| Example | Attribute (A) | | | | Example | Attribute (A) | | | |
|---|---|---|---|---|---|---|---|---|---|
| | A0 | A1 | A2 | A3 | | A0 | A1 | A2 | A3 |
| A examples | | | | | Transfer items | | | | |
| E1 | 1 | 1 | 1 | 0 | E10 | 1 | 0 | 0 | 1 |
| E2 | 1 | 0 | 1 | 0 | E11 | 1 | 0 | 0 | 0 |
| E3 | 1 | 0 | 1 | 1 | E12 | 1 | 1 | 1 | 1 |
| E4 | 1 | 1 | 0 | 1 | E13 | 0 | 0 | 1 | 0 |
| E5 | 0 | 1 | 1 | 1 | E14 | 0 | 1 | 0 | 1 |
| B examples | | | | | E15 | 0 | 0 | 1 | 1 |
| E6 | 1 | 1 | 0 | 0 | E16 | 0 | 1 | 0 | 0 |
| E7 | 0 | 1 | 1 | 0 | | | | | |
| E8 | 0 | 0 | 0 | 1 | | | | | |
| E9 | 0 | 0 | 0 | 0 | | | | | |

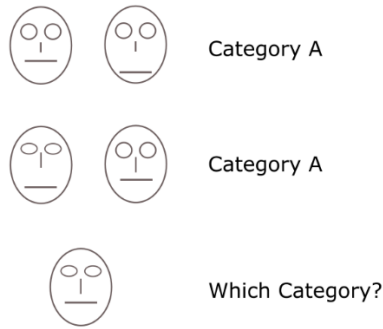

Category A

Category A

Which Category?

*Figure 10. The artificial categories with "binary" faces. The table contains binary dimensions that code facial features like nose length and eye height, the pictures show the training categories for type A and type B faces. The task is to classify novel transfer items/faces as either type A or type B.*

The basic experiment structure is presented in Figure 10. The goal is for an agent to classify a stimulus as a "type A face" or a "type B face". Unlike the two-dimensional XOR stimuli, a stimulus "5-4" face is four-dimensional, although the individual values are still binary. Examples of category A face are typically closer to having all four features turned on, while category B face has most of the four binary features turned off. The four binary values of the abstracted faces (A0 – eye height, A1 – eye separation, A2 – nose length, A3 – mouth height) provide 16 different faces, labelled E1 to E16 (see Figure 11).

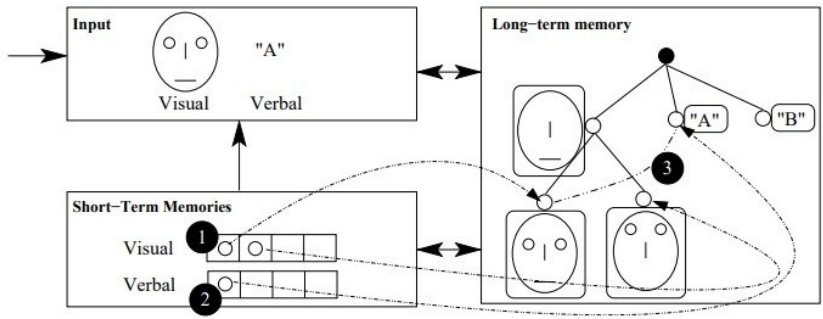

*Figure 11. CogAct's learning mechanisms in action in the five-four task: (1) the visual stimulus is sorted through LTM and a pointer to the node retrieved is formed in visual STM; (2) the verbal stimulus is sorted through LTM and pointer to the retrieved node is created in STM; and (3) when a visual pattern and a verbal pattern are stored in STM concurrently, the chunks they elicit are connected together in the LTM. From Lane and Gobet (2013)).*

The training and testing processes are similar to those in the XOR case, with the exception that CogAct automatically choses not to use rote learning as the "5-4" test stimuli are different to the training stimuli. When shown a stimuli-label pair, CogAct forms a hierarchy of visual chunks in LTM that contain the visual features of the "faces". Simultaneously, the same learning process forms nodes with verbal chunks of "A" and "B" labels (see Figure 11).  Concretely, learning in both domains comes about as the result of revising the LTM discrimination network through creating new chunks (via *discrimination*; see above) and updating the old chunks with new information (via *familiarisation*; see above). One example of discrimination would be the creation of a new "face" representation node with a "large eye height" attribute – if there was no chunk with such a face in LTM at that moment. An illustration of a chunk update would be adding the "low mouth" feature to the previously incomplete facial representation of the "low eyes plus low nose" type. When chunks from visual "face" and verbal "label"

modalities occupy the same spot in the respective STM queues, a naming link is formed

and stored into LTM. Lastly, it should also be noted that the terms "visual" and "verbal"

chunks are mere naming conventions for LTM hierarchies that represent distinct domains; the underlying symbolic nature of the patterns and the mechanisms that operate on them are exactly the same in all cases (see Figure 12 for an LTM schematic and the outputs of the model).  It is important to note that, with the five-four task, CogAct did not resort to rote learning like it did in the case of the XOR function. For instance, the model assigns label "A" to the "1 0 0 0" stimulus, despite it not being part of the training set.

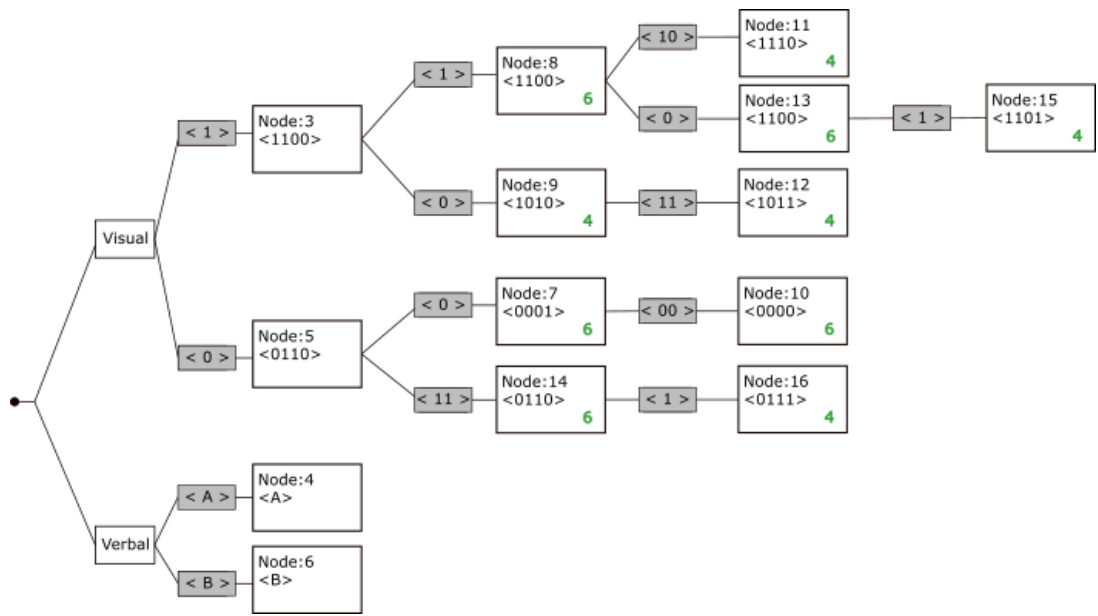

| Transfer Item / Novel Face | Ascribed Category |
|:---:|:---:|
| 1001 | A |
| 1000 | A |
| 1111 | B |
| 0010 | B |
| 0101 | A |
| 0011 | B |
| 0100 | A |

*Figure 12. CogAct's LTM and categorisation output, following an artificial "binary faces" category learning experiment. The LTM contains chunks with training binary faces in the visual node which are linked to label/ascribed category chunks in the verbal node. Green numbers denote the naming links. The table shows CogAct's categorisation performance on novel faces.*

## Category learning of artificial sequential data

Changing the order of representation of a coded photograph from *[hair colour, shirt colour, smile type, hair length]* to *[hair length, smile type, shirt colour, hair colour]* does not alter its meaning. By contrast, in language, "dog bites man" is different to "man bites dog". Similarly, different sequences of the same set of notes played on a piano can correspond to the "Happy Birthday" song, or a random piano noise. Sequential data present an additional challenge to a learning agent.

Chunking-based models derived from EPAM and CHREST have been used to simulate the acquisition of first language (Freudenthal, Pine, & Gobet, 2009; Jones et al., 2010; Tamburelli et al., 2012). These models can learn sequential data and account for important phenomena in the acquisition of vocabulary and syntactic structures. However, they still suffer from the problem of occlusion; for example, although "Liverpool" is known, the model cannot recognise the string "zLiverpool"). This is not a problem for our current model of CogAct as it has additional attentional and chunk activation mechanisms (discussed above).

We extended the original five-four task to categorising city names: "Liverpool = type A", "Manchester = type B", with different levels of occlusion: "Liverpooz", "Lizerzool", "zLiverpool", "zzzzLzverzool". Apart from the sequential and non-binary nature of the stimuli, the number of occluding letters was unlimited (e.g., "zzLiverzool", "zzzzzLiverpooz" and so on).

The procedure for the "zLiverpool" experiment remained as in the experiments above; CogAct was tasked to categorise inputs with no additional instructions provided. CogAct automatically employed its attention window and chunk activation measures. naming link / verbal chunk ("type A") from the *largest* of the retrieved visual chunks ("Liverpool")(see Figure 13 and Figure 14).

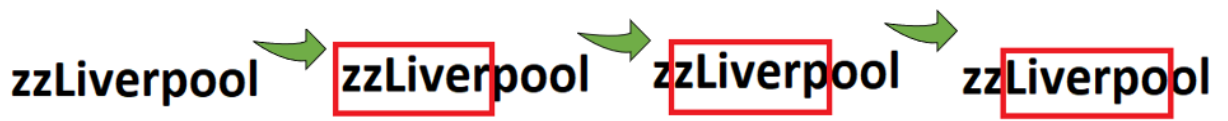

*Figure 13. CogAct's sliding attention window mechanism in action.*

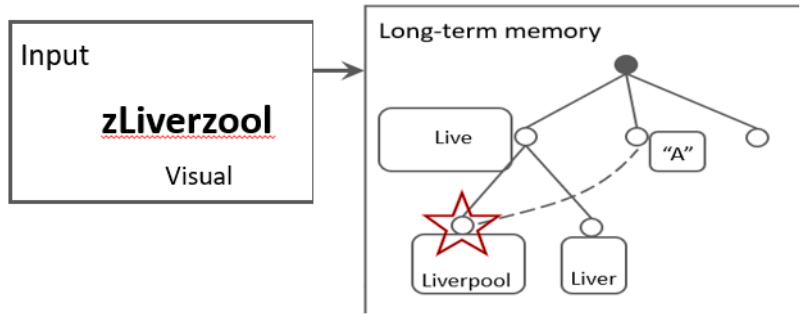

*Figure 14. CogAct's new chunk "activation" mechanism, solving a conflict between three nodes voting for zLiverzool's chunk membership. The bigger LTM chunks are rewarded, and the smaller knowledge structures are penalised.*

## Learning natural concepts in literature, music and chess

This part of the study involved learning to classify real-life categories set in the dissimilar domains of literature texts, chess openings and music scores. The procedure was the same as in the experiments above: to make CogAct learn the environmental module inputs, and then classify novel inputs. No additional instructions were provided to the model, the architecture also remained the same as in the XOR experiment and the learning of artificial categories.

The choice of literature, music and chess modalities had three primary motivations. The first was to investigate whether our cognitive model can account for significant aspects of concept formation both in linguistic (in our case, literature texts) as well as non-linguistic/quasi-linguistic (in our case, music scores and chess) complex real-life domains. This allowed us to test the "single algorithm hypothesis", which suggests that the brain uses approximately one mechanism to extract approximately one type of data structure from various types of information (Hawkins & Blakeslee, 2004; LeCun, Bengio, & Hinton, 2015; Mountcastle, 1978). Selecting three dissimilar domains also tested the sufficiency of CogAct's single mechanism – inline with the spirit of UTCs and opposed to the need, for example, of both exemplar and prototype mechanisms to account for various aspects of concept learning (Bowman et al., 2020; Machery, 2009; Murphy, 2016).

The second motivation was music specific. It involved utilizing the abstract nature of music vocabulary and semantics to reduce the limitation of many concept-formation models that are set in purely language-based domains. While verbal constructs such as "polyphony", "classical," and "baroque" are all part of the relevant musical representations, these constructs are more high level, less numerous and more isolated when compared to their literature counterparts. For example, while one could say that Homer wrote about "wrath", "Achilles," and "goddesses", there is no direct equivalent in music where one could describe Mozart composing about the "C3G4", "A4F5A5" and "D4MF4" chords and their corresponding soundwaves (see Figure 15*Figure 15* for an explanation of the music notation). This has the consequence that the music models of categorisation would be more realistic than the literature models, as the effect of semantics is minimised. Importantly, the *combination* of music, chess and literature ensures that the relatively more realistic music categorisation models are not dependant on music-specific feature detectors, but that the underlying cognitive/neural mechanisms are general.

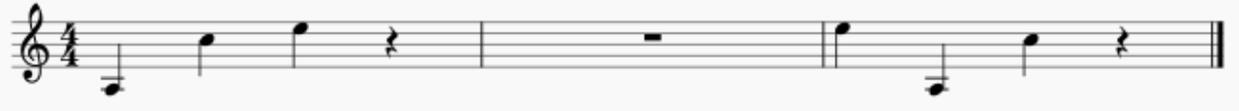

*Figure 15. The different order of writing the same three musical notes (A3, C4 and E4) alters the underlying melody: measure 1 sounds different to measure 3. In the notation used, the letters denote the musical notes, while the numbers denote the octaves – e.g., A3 refers to the note A in the 3rd octave.*

The third motivation was to test if the performance of our domain-general CogAct was comparable to the previous chess-specific categorisation approach within the chunking theory and CHREST (Lane & Gobet, 2012c).

We trained the model on unabridged works by various authors, composers, and a chess openings database. We tested categorisation on previously unseen pieces produced by the same authors and composers, as well as unseen chess openings. CogAct was never told which domain formed its input, with both training and testing being done with raw data only.

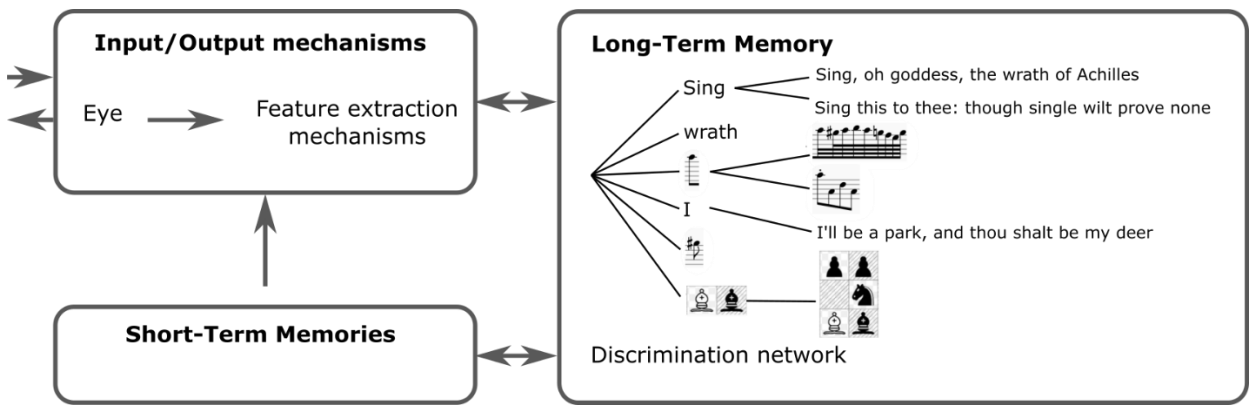

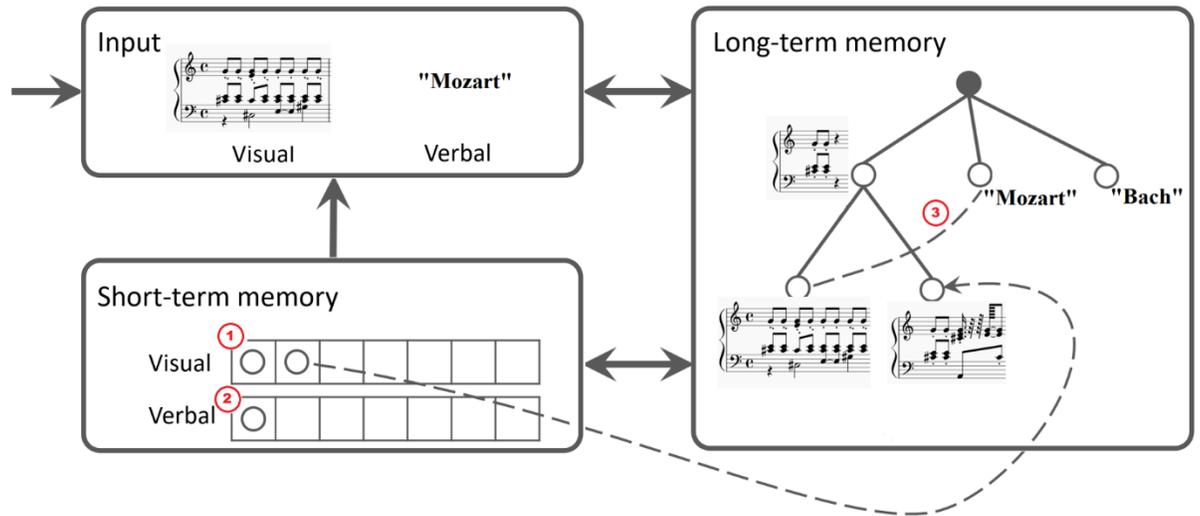

*Figure 16. CogAct's learning mechanisms in action: (1) the visual stimulus is sorted through LTM and a pointer to the node retrieved is formed in visual STM; (2) the verbal stimulus is sorted through LTM and pointer to the retrieved node is created in STM; and (3) when a visual pattern and a verbal pattern are stored in STM concurrently, the chunks they elicit are connected together in the LTM.*

The basic CogAct structure is presented in Figure 16. For example, when shown a music pieces, CogAct forms a hierarchy of visual chunks in LTM that contain the visual features of the chords and notes. Concomitantly, the same learning process forms nodes with verbal chunks of the labels such as "Mozart" and "Bach". Concretely, learning in both domains comes about as the result of revising the LTM discrimination network through *creating* new chunks and *updating* the old chunks with new information. One example of the former process would be the creation of a new chord representation node with a "A2E3" chord – if there was no chunk with such a chord in LTM at that moment. An illustration of a chunk update would be adding the "F2" note to the already known "A2E3" part of the sequence. When chunks from visual "chords" and verbal "label" modalities occupy the same spot in the respective STM queues, a naming link is formed and stored into LTM. Lastly, it should also be noted that the terms "visual" and "verbal" chunks are mere naming conventions for LTM hierarchies that represent distinct domains; the underlying symbolic nature of the patterns and the mechanisms that operate on them are exactly the same in all cases. The literature and chess experiments were identical to the above, but with input being text instead of music scores.

### Method

*Training and testing: Literature.*

The training data for the literature modality were as follows. For Shakespeare, CogAct was given 140 *Sonnets*, *Romeo and Juliet* and an excerpt of *Midsummer Night's Dream*. The Homer training set had the first four chapters of *The Iliad* (translated by Samuel Butler). For Dickens, a large excerpt from *David Copperfield* was used. For Chaucer, it was *Troilus and Criseyde*.  For Walter Scott, the "reading set" included excerpts from *Ivanhoe* and *Rob Roy*. Lastly, the Joyce sample contained the first four chapters of *Ulysses*. For every author, there was 300Kb of text in total.

With respect to testing, the Shakespeare tests included the remaining *Sonnets*, *The Passionate Pilgrim*, *Venus and Adonis*, *The Comedy of Errors*, *Pericles Prince of Tyre,* and *Macbeth and Hamlet*. The Homer tests contained excerpts from *The Odyssey* and the ending chapter of *The Iliad*. The Chaucer tests included excerpts from *Canterbury Tales*, *Book of Duchesse*, *The House of Fame,* and *The Parliament of Fowles*. The Dickens test was comprised of excerpts from *The Great Expectations*,

*Little Dorrit*, *Oliver Twist, The Pickwick Papers*, *A Christmas Carol,* and *Tale of Two Cities*. The Walter Scott test category had excerpts from *The Lay of The Last Minstrel*, *The Black Dwarf*, *Marmion,* and *Talisman*. The final chapter of *Ulysses*, excerpts from *Finnegans Wake,* and *A Portrait of the Artist as a Young Man* formed the Joyce test. Altogether, there were 60 novel literature pieces (10 per each author).

*Training and testing: Music.*

The music modality training was comprised of the following. For Bach, CogAct was given 63 pieces from Bach's *Well-Tempered Clavier (WTC)*. Mozart's *Piano Sonatas No.1-6* and *No.8-12*; Beethoven's *Piano Sonatas No.1-7*; and 14 *Etudes* and 18 *Nocturnes* by Chopin formed the rest of the training. As with literature, each training category contained approximately 300Kb of text, this time generated from MIDI[5] files.

The music test dataset contained 15 Bach *WTC* pieces; 9 *Sonata* pieces, Mozart's *A piece for piano K176*, *Adagio in B flat*, *Fantasia in C*, *Fantasia in D*, *Adagio K617* and *Rondo in C*; Beethoven's 15 *Sonata* movements; Chopin's 4 *Waltzes*, 3 *Sonatas, Ballad Op.32*, a *Nocturne*, two *Etudes*, and one *Polonaise*. Altogether, every composer had 15 novel pieces for CogAct to categorise. Like with literature, there were 60 novel music pieces for CogAct to categorise.

---

[5] MIDI (Musical Instrument Digital Interface) is a standard protocol making it possible to exchange musical information between computers, synthesizers, and musical instruments. Unlike the widely known MP3 files, the MIDI file format does not represent sound waveforms. Instead, MIDI contains the information on the sequence and duration of the played notes.

All pieces were transposed to C-major/A-minor key. Full chord complexity and polyphony was preserved, but timings were not kept in the text conversions. The pattern primitives for the text modality were words. In the music case, the pattern primitives for the music modality were note/chord structures that occupied one time step in a given sequence (see Figure 17).

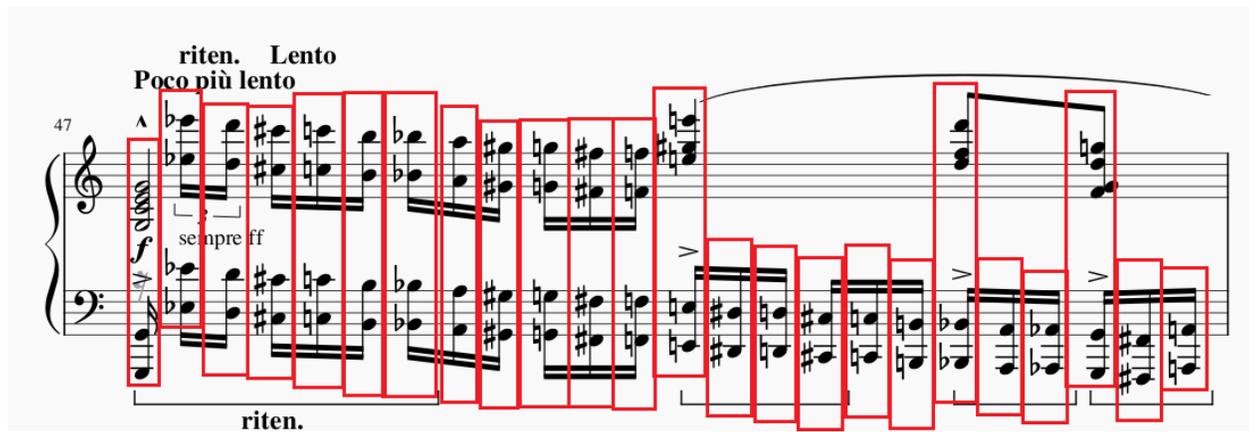

*Figure 17. An excerpt from Chopin's Op.48 No1. From the perspective of CogAct, each vertical frame represents a single pattern primitive, equivalent to a word in a text sentence.*

*Training and testing: Chess.*

The chess modality training involved two categories – French and Sicilian openings – with CogAct receiving approximately 21Kb of chess positions data for each category. This means that the chess category contained 42Kb of text, as opposed to the approximately 300Kb with literature and music. See Figure 18 for the examples of these chess positions.

The chess test dataset contained (previously unseen) 30 French and 27 Sicilian openings, making for a total of 57 chess categorisations.

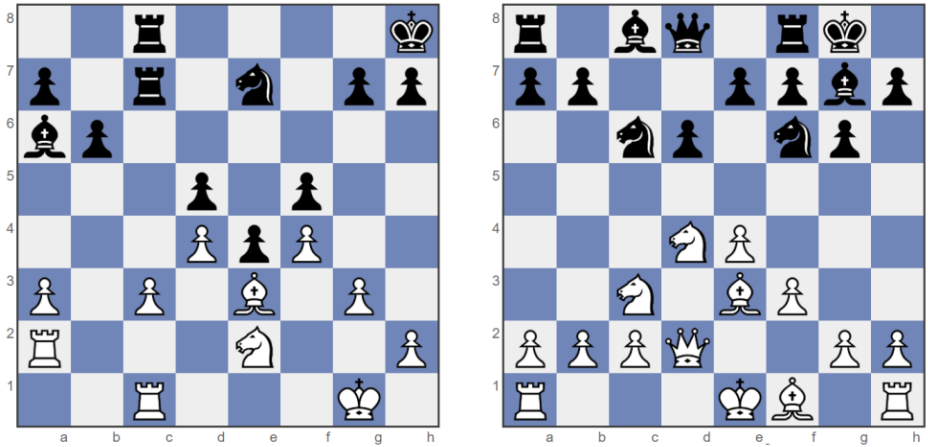

*Figure 18. Examples of a French opening (on the left) and a Sicilian opening (on the right).*

*Procedure.*

The internal parameters of the model were fixed for the entire duration of the experiment. In particular, the STM size was set to 5 chunks; the maximum size of the attention window was set to 20 words or 1 measure; the likelihood of forming a chunk was set to 1; the simulated time needed to create a new chunk was set to 10 seconds; the time needed to update a chunk was set to 2 seconds. The model was trained until no learning was possible. Chess, Music and literature patterns were assigned to the visual modality, while author/composer/chess opening names were assigned to the verbal modality.

Following the conclusions of Gobet and Lane (2012), the training samples were split into 20-word phrases (for text), 2 measures (for music) and 1 row/ 8 symbols (for chess) to avoid forming overly large chunks. The order of the training samples was randomised. No words/notes/chess positions were removed from either training or testing texts/music/chess scores.

*Transparency and openness.*

We report all data, manipulations, analysis, and source code. The program code was written in Python version 3.7.5 and Tensorflow version 2.0.0. The used Python modules were as follows: Numpy version 1.17.4, RE version 2.2.1, JSON version 2.0.9, CSV version 1.0. See https://github.com/Voskod/CogAct for all program source code that was used in the current article. The stimuli data could not be put online due to potential copyright issues, please email the authors for access. The study's design and its analysis were not pre-registered. The ethics committee declared this study exempt as it reports human simulations and no human data were collected. This study was not preregistered.

### Results

By the end of training, CogAct's LTM developed around 34,000 chunks that encoded the literature modality, 22,000 chunks that encoded music and around 3,000 chunks encoding chess. Overall, CogAct was able to learn and apply new concepts in the complex real-world domains of literature, music and chess. It required no ad hoc additions to the fundamental architecture in order to deal with domain-specific nuances. The detailed breakdown of the literature and music categorisation performance is presented in Table 1; chess position categorisation performance was around 85%.

CogAct's categorisation performance was substantially above chance, CogAct correctly categorised 41 out of 60 literature works (with the random chance baseline being 10 correct guesses); 42 out of 60 music scores (with the same baseline being 15) and 53 out of 61 chess positions (against the baseline of 31).

*Table 1. CogAct's categorisation of 60 music works and 60 literature pieces. Red colour signifies weak activation and darker shades of green signify stronger activation. Numbers in bold signify its highest confidence score on a given test.*

| | File | Chaucer | Dickens | Homer | Joyce | Scott | Shakespeare | Correct |
|---|---|---|---|---|---|---|---|---|
| **Chaucer** | Book Of The Duchesse_P1 | **0.46** | 0.10 | 0.28 | 0.07 | 0.04 | 0.04 | TRUE |
| | Book Of The Duchesse_P2 | **0.45** | 0.09 | 0.18 | 0.07 | 0.11 | 0.11 | TRUE |
| | Canterbury_Tale1 | **0.59** | 0.08 | 0.14 | 0.05 | 0.02 | 0.13 | TRUE |
| | Canterbury_Tale2 | **0.51** | 0.11 | 0.18 | 0.04 | 0.06 | 0.10 | TRUE |
| | Canterbury_Tale3 | **0.50** | 0.10 | 0.14 | 0.05 | 0.08 | 0.12 | TRUE |
| | The House Of Fame_P1 | **0.47** | 0.08 | 0.25 | 0.09 | 0.05 | 0.07 | TRUE |
| | The House Of Fame_P2 | **0.53** | 0.08 | 0.14 | 0.04 | 0.04 | 0.16 | TRUE |
| | The House Of Fame_P3 | **0.45** | 0.12 | 0.10 | 0.05 | 0.12 | 0.16 | TRUE |
| | The Parliament Of Fowles_P1 | **0.49** | 0.14 | 0.09 | 0.06 | 0.08 | 0.15 | TRUE |
| | The Parliament Of Fowles_P2 | **0.58** | 0.05 | 0.13 | 0.04 | 0.03 | 0.17 | TRUE |
| **Dickens** | Christmas | 0.05 | **0.30** | 0.15 | 0.16 | 0.18 | 0.16 | TRUE |
| | Great_Expectations_P1 | 0.10 | **0.29** | 0.17 | 0.18 | 0.12 | 0.14 | TRUE |
| | Great_Expectations_P2 | 0.07 | **0.33** | 0.14 | 0.14 | 0.15 | 0.16 | TRUE |
| | Great_Expectations_P3 | 0.05 | **0.34** | 0.20 | 0.14 | 0.11 | 0.17 | TRUE |
| | Little_Dorrit_P1 | 0.04 | 0.22 | **0.23** | 0.14 | 0.18 | 0.20 | FALSE |
| | Little_Dorrit_P2 | 0.06 | **0.21** | 0.19 | 0.17 | 0.22 | 0.15 | FALSE |
| | Oliver_Twist_Ch1_Ch2 | 0.06 | **0.22** | 0.17 | 0.16 | 0.21 | 0.18 | TRUE |
| | Oliver_Twist_Ch3_Ch4 | 0.05 | 0.23 | 0.14 | 0.14 | **0.24** | 0.16 | FALSE |
| | Pickwick_Papers_Ch2 | 0.06 | **0.25** | 0.15 | 0.17 | 0.18 | 0.19 | TRUE |
| | Tale Of 2 Cities | 0.06 | **0.26** | 0.21 | 0.09 | 0.21 | 0.18 | TRUE |
| **Homer** | Iliad end chapter | 0.05 | 0.18 | **0.44** | 0.08 | 0.11 | 0.14 | TRUE |
| | Odyssey_1 | 0.15 | **0.30** | 0.07 | 0.10 | | 0.16 | FALSE |
| | Odyssey_2 | 0.16 | 0.27 | **0.27** | 0.05 | 0.13 | 0.13 | TRUE |
| | Odyssey_3 | 0.05 | 0.23 | **0.33** | 0.10 | 0.16 | 0.13 | TRUE |
| | Odyssey_4 | 0.05 | 0.20 | **0.29** | 0.13 | 0.19 | 0.15 | TRUE |
| | Odyssey_5 | 0.11 | 0.15 | **0.28** | 0.16 | 0.19 | 0.10 | TRUE |
| | Odyssey_6 | 0.08 | 0.16 | **0.33** | 0.11 | 0.16 | 0.17 | TRUE |
| | Odyssey_7 | 0.10 | 0.21 | **0.33** | 0.06 | 0.12 | 0.18 | TRUE |
| | Odyssey_8 | 0.07 | 0.17 | 0.24 | 0.10 | **0.25** | 0.17 | FALSE |
| | Odyssey_9 | 0.07 | 0.21 | **0.28** | 0.11 | 0.15 | 0.18 | TRUE |
| **Joyce** | A_Portrait_Of_The_Artist _1 | 0.05 | **0.25** | 0.14 | 0.24 | 0.20 | 0.11 | FALSE |
| | A_Portrait_Of_The_Artist _2 | 0.04 | 0.20 | 0.14 | **0.24** | 0.21 | 0.16 | TRUE |
| | A_Portrait_Of_The_Artist _3 | 0.02 | 0.23 | 0.20 | **0.23** | 0.10 | 0.21 | TRUE |
| | Finnegans_Wake_1 | 0.12 | 0.12 | 0.19 | 0.18 | 0.10 | **0.29** | FALSE |
| | Finnegans_Wake_2 | 0.13 | 0.16 | 0.18 | 0.09 | 0.16 | **0.29** | FALSE |
| | Finnegans_Wake_3 | 0.11 | 0.21 | **0.23** | 0.11 | 0.12 | 0.22 | FALSE |
| | Finnegans_Wake_4 | 0.22 | 0.15 | 0.17 | 0.10 | 0.10 | **0.27** | FALSE |
| | Finnegans_Wake_5 | 0.07 | **0.31** | 0.19 | 0.15 | 0.12 | 0.16 | FALSE |
| | Ulysses_Ending_P1 | 0.09 | 0.22 | 0.14 | 0.08 | **0.23** | 0.19 | FALSE |
| | Ulysses_Ending_P2 | 0.10 | **0.29** | 0.19 | 0.18 | 0.13 | 0.12 | FALSE |
| **Scott** | Black Dwarf_P1 | 0.08 | 0.19 | 0.15 | 0.12 | **0.28** | 0.18 | TRUE |
| | Black Dwarf_P2 | 0.10 | 0.14 | 0.18 | 0.14 | **0.23** | 0.22 | TRUE |
| | Marmion1 | 0.08 | 0.20 | **0.24** | 0.09 | 0.19 | 0.20 | FALSE |
| | Marmion2 | 0.05 | 0.22 | 0.16 | 0.13 | 0.12 | **0.32** | FALSE |
| | Talisman_P1 | 0.06 | **0.22** | 0.17 | 0.14 | 0.20 | 0.21 | FALSE |
| | Talisman_P2 | 0.05 | 0.22 | 0.21 | 0.08 | **0.26** | 0.17 | TRUE |
| | Talisman_P3 | 0.05 | 0.21 | 0.20 | 0.14 | **0.25** | 0.16 | TRUE |
| | Talisman_P4 | 0.11 | 0.21 | 0.21 | 0.10 | 0.16 | **0.22** | FALSE |
| | Talisman_P5 | 0.11 | 0.17 | **0.26** | 0.11 | 0.17 | 0.18 | FALSE |
| | The Lay Of The Last Minstrel | 0.04 | 0.13 | 0.11 | 0.15 | 0.17 | **0.39** | FALSE |
| **Shakespeare** | Comedy Of Errors | 0.05 | 0.20 | **0.26** | 0.09 | 0.14 | 0.26 | FALSE |
| | Hamlet_P1 | 0.07 | 0.18 | 0.17 | 0.13 | 0.17 | **0.28** | TRUE |
| | Hamlet_P2 | 0.13 | 0.17 | 0.15 | 0.08 | 0.13 | **0.34** | TRUE |
| | Macbeth_P1 | 0.08 | 0.19 | 0.20 | 0.11 | 0.13 | **0.29** | TRUE |
| | Macbeth_P2 | 0.04 | 0.11 | 0.18 | 0.15 | 0.20 | **0.33** | TRUE |
| | Macbeth_P3 | 0.04 | 0.11 | **0.30** | 0.13 | 0.15 | 0.26 | FALSE |
| | Pericles | 0.09 | 0.14 | 0.19 | 0.11 | 0.15 | **0.31** | TRUE |
| | Pilgrim | 0.13 | 0.17 | 0.14 | 0.12 | 0.03 | **0.40** | TRUE |
| | Sonnets_2 | 0.17 | 0.04 | 0.08 | 0.04 | 0.04 | **0.63** | TRUE |
| | Venus | 0.12 | 0.23 | 0.14 | 0.07 | 0.12 | **0.32** | TRUE |

| | File | Bach | Beethoven | Chopin | Mozart | Correct |
|---|---|---|---|---|---|---|
| **Bach** | WTCII09A | **0.56** | 0.26 | 0.10 | 0.09 | TRUE |
| | Wtcii09b | **0.59** | 0.26 | 0.11 | 0.04 | TRUE |
| | Wtcii10a | **0.57** | 0.16 | 0.11 | 0.15 | TRUE |
| | Wtcii10b | **0.52** | 0.23 | 0.10 | 0.15 | TRUE |
| | Wtcii11a | **0.57** | 0.31 | 0.09 | 0.03 | TRUE |
| | Wtcii11b | **0.59** | 0.27 | 0.07 | 0.08 | TRUE |
| | WTCII12A | **0.41** | 0.24 | 0.14 | 0.20 | TRUE |
| | Wtcii12b | **0.56** | 0.17 | 0.09 | 0.18 | TRUE |
| | WTCII13A | **0.47** | 0.17 | 0.17 | 0.19 | TRUE |
| | Wtcii13b | **0.53** | 0.23 | 0.07 | 0.17 | TRUE |
| | WTCII21A | **0.51** | 0.23 | 0.11 | 0.15 | TRUE |
| | Wtcii21b | **0.50** | 0.28 | 0.06 | 0.16 | TRUE |
| | WTCII22A | **0.60** | 0.18 | 0.10 | 0.13 | TRUE |
| | Wtcii22b | **0.58** | 0.21 | 0.11 | 0.10 | TRUE |
| | WTCII23A | **0.55** | 0.28 | 0.09 | 0.08 | TRUE |
| **Beethoven** | Piano Sonata N08 Op13 1mov | 0.31 | **0.44** | 0.14 | 0.10 | TRUE |
| | Piano Sonata N08 Op13 3mov | **0.46** | 0.32 | 0.12 | 0.11 | FALSE |
| | Piano Sonata N09_1 | 0.33 | **0.50** | 0.10 | 0.06 | TRUE |
| | Piano Sonata N09_2 | 0.34 | **0.51** | 0.09 | 0.06 | TRUE |
| | Piano Sonata N10 1mov | **0.41** | **0.41** | 0.08 | 0.09 | TRUE |
| | Piano Sonata N10 2mov | 0.30 | **0.50** | 0.09 | 0.11 | TRUE |
| | Piano Sonata N14 Op27 1mov ''Moonlight'' | 0.41 | **0.48** | 0.07 | 0.03 | TRUE |
| | Piano Sonata N14 Op27 3mov ''Moonlight'' | 0.34 | **0.38** | 0.19 | 0.09 | TRUE |
| | Piano Sonata N15_1 | 0.16 | **0.50** | 0.16 | 0.18 | TRUE |
| | Piano Sonata N15_2 | 0.32 | **0.43** | 0.13 | 0.13 | TRUE |
| | Piano Sonata N17 Tempestat_1 | 0.35 | **0.37** | 0.14 | 0.14 | TRUE |
| | Piano Sonata N17 Tempestat_3 | 0.37 | **0.44** | 0.15 | 0.05 | TRUE |
| | Piano Sonata N18 The Hunt_1 | 0.28 | **0.41** | 0.18 | 0.13 | TRUE |
| | Piano Sonata N18 The Hunt_3 | 0.34 | **0.46** | 0.11 | 0.09 | TRUE |
| | Piano Sonata N21 1movWaldstein | 0.30 | **0.47** | 0.12 | 0.11 | TRUE |
| **Chopin** | 2 Polonaises Op26 N1 | 0.00 | 0.14 | 0.30 | **0.55** | FALSE |
| | Ballad Op32 | 0.11 | 0.09 | **0.46** | 0.35 | TRUE |
| | Etude Op10 N01 | **0.35** | 0.27 | 0.29 | 0.09 | FALSE |
| | Grande Waltz Op18 | 0.02 | 0.12 | 0.18 | **0.68** | FALSE |
| | Nocturne Op. 9 No. 1 | 0.34 | **0.42** | 0.19 | 0.05 | FALSE |
| | Prelude N15 Op28 ''Raindrop'' | 0.05 | 0.23 | 0.30 | **0.42** | FALSE |
| | Scherzo N1 Op20 | 0.08 | 0.12 | 0.31 | **0.49** | FALSE |
| | Scherzo Op31 | 0.05 | 0.11 | 0.29 | **0.55** | FALSE |
| | Sonata Op35 N2 | 0.19 | 0.29 | **0.35** | 0.17 | TRUE |
| | Sonata Op35 N3 | 0.12 | 0.16 | 0.35 | **0.37** | FALSE |
| | Sonata Op35 N4 | 0.03 | 0.30 | 0.29 | **0.38** | FALSE |
| | TroisNouvellesEtudes_Chopin_N1 | **0.58** | 0.23 | 0.13 | 0.05 | FALSE |
| | Waltz In Ab (Op.Posth., Brown21) | 0.04 | 0.04 | 0.40 | **0.53** | FALSE |
| | Waltz Op64 N1 | 0.03 | 0.10 | **0.54** | 0.33 | FALSE |
| | Waltz Op70 N2 | 0.18 | 0.16 | 0.32 | **0.34** | FALSE |
| **Mozart** | A Piece For Piano K176 | **0.60** | 0.16 | 0.03 | 0.21 | FALSE |
| | Adagio In B Flat | 0.12 | 0.04 | 0.29 | **0.55** | TRUE |
| | Fantasia In C K.475 | 0.14 | 0.20 | 0.28 | **0.38** | TRUE |
| | Fantasia In D, K397 | 0.05 | 0.06 | 0.31 | **0.58** | TRUE |
| | K309 Piano Sonata N10 1mov | **0.44** | 0.24 | 0.11 | 0.22 | FALSE |
| | K309 Piano Sonata N10 2mov | 0.07 | 0.11 | 0.30 | **0.53** | TRUE |
| | K331 Piano Sonata N11 | 0.10 | 0.14 | 0.25 | **0.52** | TRUE |
| | K332 Piano Sonata N12 1mov | 0.04 | 0.04 | 0.18 | **0.75** | TRUE |
| | K333 Piano Sonata N13 1mov | 0.08 | 0.12 | 0.31 | **0.49** | TRUE |
| | K333 Piano Sonata N13 3mov | 0.07 | 0.08 | 0.30 | **0.55** | TRUE |
| | K401 Fuga In G Minor | **0.53** | 0.27 | 0.08 | 0.12 | FALSE |
| | K570 Piano Sonata 1mov | 0.04 | 0.06 | 0.23 | **0.66** | TRUE |
| | K570 Piano Sonata 3mov | 0.06 | 0.09 | 0.32 | **0.53** | TRUE |
| | K617 Adagio | 0.30 | **0.46** | 0.07 | 0.17 | FALSE |
| | Rondo In C | 0.30 | **0.43** | 0.11 | 0.16 | FALSE |

**Discussion**

The CogAct modelling has several key strengths and makes important contributions. Firstly, the computational methodology allowed for rigorous and objective investigation of the fundamental learning mechanisms implicated in human concept formation. Secondly, the model moved away from relying on hand engineered features/dimensions while learning from complex real-life raw data from multiple modalities. Third, CogAct demonstrated its ability to adapt to wide range of categories without bootstrapping to pre-built knowledge structures, learning from raw data in all cases. Another strength of this study is its intuitive realism with regard to its parsimony with training data – only a fraction of Homer's *Iliad* and several Mozart's *Sonatas* were needed to learn generalisable concepts of Homer and Mozart respectively. At this point we should note that research into chess expertise has shown that over 300,000 LTM domain-specific chunks are needed for experts to perform true to their name (Gobet & Simon, 1998; Richman et al., 1996). The LTM size of the current model was way below that, which may possibly explain some of its performance.

It is also noteworthy that CogAct managed to learn the chess position categorisation without chess specific hand-crafted feature detectors. CogAct's accuracy in differentiating the French and Sicilian openings is similar to what was reported with CHREST (which relied on hand-crafted chess features designed by an International Master level player) – around 85% for both CHREST and CogAct (Lane & Gobet, 2012c).

One potential theoretical weakness is that the author/composer categories were a-priori labelled and predetermined, with learning being supervised and closed to unsupervised clustering of input examples. There are two ways of answering this criticism. Firstly, the CogAct architecture has unsupervised learning at its core: the automatic clustering of patterns into chunks is independent of a-priori labels. Secondly, human readers and musicians do tend to know the name of the author/composer that they are studying, thus largely matching the labelled stimuli approach of this concept formation study.

The last point forms a segue to the next section, where the task of modelling human music expert's concept spaces further grounds the psychological mechanisms of CogAct.

## Modelling Human Subjective Concept Space

A good psychological theory satisfies multiple constraints to limit our search in the otherwise infinite space of psychological candidate theories. In the previous sections, we demonstrated how CogAct incorporates major cognitive mechanisms while also adhering to the "single algorithm hypothesis".

The current section aims to further ground our CogAct model with simulations of human subjective evaluation of complex natural categories in music. Consequently, the remainder of the paper is split into three studies. Firstly, we establish the human "training data" in the shape of participants' experience with the tested categories and we measure human categorisation performance in the form of participants' predictions on novel stimuli. Secondly, we then develop a bespoke CogAct model for each of the participants by training it on the same training data and testing it on the same novel stimuli as the corresponding human participant. In the third part, we present a deep learning model as a baseline comparison.

One traditional difficulty with modelling human performance is the problem of prior knowledge. For example, even if a psychological model were to completely replicate the human performance on a particular test, the question of what training data were used for learning would also need to be answered. How much training data did a human need to reach the same level of performance as the model? With many natural domains, the question of prior knowledge may be answered only in very approximate terms (e.g., Battleday et al., 2020; Dyck et al., 2021; Elsayed et al., 2018; Firestone, 2020; Mnih et al., 2013). As an example, consider a hypothetical psychological model of visual categorisation. Modelling categorisation of natural visual categories such as "a table" inevitably leads to questions of how many tables and what kind of tables were seen by the human and to what extent this training set was matched by the model. The traditional way of circumventing this issue was either to ignore it or to use artificial concepts (Vanpaemel & Storms, 2008). The downside of that approach was that the artificial stimuli tended to be very simple with the possibility of being contaminated by items from a natural category (e.g., an abstract stimulus in the form of a geometrical 3D angle shape being associated with a coat hanger).

We chose music (in the form of classical piano scores) to be our experimental modality. This allowed for tractable investigation of participants' training data all while maintaining the ecological validity of the natural category stimuli in a realistic highly complex domain. While it is intractable to ask a participant how many tables she saw in her lifetime, it is a reasonable question to ask how many Beethoven pieces she attentively studied. The second important reason for choosing music, as was discussed above, involved utilizing the abstract nature of its vocabulary and semantics so as to reduce the impact of oversimplification inherent in the models that are set in linguistic domains.

An additional strength of the current study was its aim to capture participants' *subjective* concept spaces. For example, if a participant's listening history of Mozart and Bach consisted of *Fantasia in D* and the *WTC Fugue 24* respectively, the training data for modelling would be *Fantasia in D* and the *WTC Fugue 24*. This would imply that the models would be *of* that participant – as opposed to the commonly used approach of modelling the "average" person. Accordingly, the same participant may categorise a small excerpt of the (novel to her) *Sonata K357* piece as Mozart, and the model would generate its predictions of the same piece, allowing for direct comparisons with the human benchmark. Given that the aim of the experiment was to model individual categorisation performance based on an individual's unique past experience, the number of participants was small.

Consequently, the current study was split into two parts. The first part was tasked with procuring the human data, both with regard to the music history and the categorisation performance on novel pieces. The focus of the music history was on the specific composers that would be used in the experiment, with participants recalling both their listening and their performance repertoire for the composers chosen by the experimenter. The corresponding categorisation performance was measured on music pieces that were not reported as part of the participant's individual musical history.

The second and third parts involved training and testing the CogAct models and baseline ANN models with the data generated in the first study.

## Method

### *Participants.*

The participants (2 females, 4 males; age range 18 – 30 years) consisted of six music conservatory students specialising in classical piano performance. The participants were recruited using the opportunity sampling method. The study was approved by the relevant institutional ethics committee. This study was not preregistered.

### *Design and materials.*

The main instrument of the experiment was a structured interview split over two sessions. Both sessions took place over Skype due to Covid-19 and were audio-recorded. MuseScore3 program was used to playback the MIDI audio files. The MIDI audio files came from the personal music library of the first author. All test music pieces were transposed to A-minor and C-major in MuseScore3. The full list of works of each composer was obtained either from the researcher's personal music library or the *http://kunstderfuge.com* website.

### *Procedure.*

An advertisement for the study was emailed to music conservatory students. The six participants were given information and consent forms. Participants agreed to complete two structured interview sessions (carried out over Skype), with the second given approximately two weeks after the first. The two sessions lasted around one hour each. The participants were paid 20 UK Pounds for taking part.

Once participants had agreed to take part in the study and signed the consent forms, the researcher contacted them via Skype. Standardised assessments were used to gain a picture of the participants' experience in music.

During the first session, the participants were asked to report their musical history – i.e., which pieces by which classical composers they have studied and attentively listened in the past. During the second session, the participants were given audio MIDI segments and asked to categorise them as belonging to one of the classical composers mentioned during the first session. Crucially, these segments came from works that the participants did not report they had studied/listened during the first session. The test audio MIDI segments were approximately one minute in duration.

The interview schedule was the following:

Session 1:

1. Given a particular composer, which classical piano pieces do you recall studying and/or listening? State the name of the piece and the composer. Use the full list of works for each composer to help your recall of previously studied musical pieces.

Session 2:

1. Which of the listed composers would you say has written the following excerpt of music?

2.  Please say if you recognise/know the piece and it will be skipped.

3.  Please rate your confidence by a score out of 10 (e.g., 9/10). Please give your second-choice composer if your confidence rating is below 10/10.

4.  Please think aloud or explain your reasoning for giving your answer where you can.

### Design of simulations.

For each simulated participant, the CogAct and ANN models were trained on the same pieces as those reported in the corresponding human participant's musical history. The models were also tested on the same novel music files as the corresponding participant. The human data were directly compared to the predictions of the models. The detailed scores of the ANN models can be found in the supplementary online material section at https://github.com/Voskod/CogAct.

Five progressively less stringent comparison binary metrics were developed to aid comparison: *Identical*, *Both match*, *Tops match*, *One matches top*, and *Single match*. The metrics had the following methods of scoring:

1.  The *Identical* metric scored "1", where human and model predictions were identical, including the order of the choices; for example, (Bach, Mozart) for the human and (Bach, Mozart) for the model.

2.  The *Both match* metric scored "1", where the model prediction choices were the same as the human choices, disregarding the order effect. For example, a stimulus that was categorised as (Bach, Mozart) by a human, and (Mozart, Bach) by the model would score a "1". With some of the composers being closer to each other in the subjective conceptual space (e.g. Haydn and Mozart), the goal was to test if the models were able to capture such inter-concept relationships.

3. The *Tops match* metric scored "1" where the highest model prediction matched the most confident human prediction. An example is: (Bach, Mozart) for the human prediction, and (Bach, Haydn) for the model prediction.

4. The *One matches top* metric scored "1" where any of the two top model's predictions matched the top human prediction. An example would include the human prediction (Bach, Mozart) and the model prediction (Haydn, Bach).

5. The *Single match* metric scored "1" where at least one of the two top model predictions matched at least one of the human predictions. For example, a stimulus that was categorised as (Bach, Mozart) by a human and (Haydn, Mozart) by the model would score a "1". The motivation for this test was to provide a more inclusive version of the *Both match* metric that would test if at least one of the model's predictions could be deemed plausible according to the corresponding participant's subjective rating.

The *Identical* metric was the most stringent and the *Single match* metric was the least stringent in the metric set.

The data for each participant are presented first, followed by modelling and a statistical discussion of the results.

# Results

### Participant AB.

Composer-specific piano music history (items with a star indicate answers where the participant was less than certain): Bach – the entire *Well Tempered Clavier (WTC)*, the entire *Goldberg Variations*; Beethoven – all 32 *Piano Sonatas* and all 5 *Piano Concertos*; Mozart – *Piano Concertos No5* and *No18\**, *Piano Sonatas No1-No8\**, *Fantasia in C*; Schubert – all of the *Impromptus*, *Piano Sonatas No18-No21\**.

The categorisation scores of novel music pieces can be seen in Table 2. (In this table, the red colour signifies zero activation and darker shades of green signify stronger activation.)

*Table 2. Categorisation of 20 novel music pieces by AB and her corresponding CogAct model. The scores in bold denote the highest confidence score on a given test.*

| | File | CogAct model | | | | Human | | | |
|---|---|---|---|---|---|---|---|---|---|
| | | Bach | Beethoven | Mozart | Schubert | Bach | Beethoven | Mozart | Schubert |
| Bach | Bwv0525 Sonate En Trio N1_39-100sec | **0.5** | **0.5** | 0 | 0 | **0.7** | 0 | 0.3 | 0 |
| | Bwv0526 Sonate En Trio N2_39-100sec | **0.5** | **0.5** | 0 | 0 | **1** | 0 | 0 | 0 |
| | Bwv0527 Sonate En Trio N3_39-100sec | **0.6** | 0.4 | 0 | 0 | **1** | 0 | 0 | 0 |
| | Bwv0528 Sonate En Trio N4_39-100sec | **0.6** | 0.4 | 0 | 0 | **0.7** | 0.3 | 0 | 0 |
| | Bwv0529 Sonate En Trio N5_39-100sec | **0.5** | **0.5** | 0 | 0 | **1** | 0 | 0 | 0 |
| Beethoven | Anh06 Rondo_39-100sec | 0 | **0.6** | 0 | 0.4 | 0 | 0 | **0.7** | 0.3 |
| | Anh08Nb1 Gavotte 4 Hands_39-100sec | 0 | **0.7** | 0.3 | 0 | 0 | 0.3 | **0.7** | 0 |
| | Bagatella Op33 N1_39-100sec | 0 | **0.6** | 0.4 | 0 | 0 | 0 | **0.8** | 0.2 |
| | Bagatella Op33 N2_39-100sec | 0.3 | **0.7** | 0 | 0 | 0 | 0.2 | 0 | **0.8** |
| | Bagatella Op33 N3_39-100sec | 0.3 | **0.7** | 0 | 0 | 0 | 0.3 | **0.7** | 0 |
| Mozart | Adagio In B Flat_38-98sec | 0 | 0.4 | **0.6** | 0 | 0 | **0.4** | 0.2 | **0.4** |
| | Another Piece For Piano,K.176_38-98sec | 0 | **0.5** | **0.5** | 0 | 0 | **0.5** | **0.5** | 0 |
| | K19d Piano Sonata Duet_38-98sec | 0 | **0.5** | **0.5** | 0 | 0 | 0.2 | **0.8** | 0 |
| | K401 Fuga In G Minor _38-98sec | 0.4 | **0.6** | 0 | 0 | **1** | 0 | 0 | 0 |
| | K497 Piano Sonata 4 Hands_38-98sec | 0 | **0.6** | 0.4 | 0 | 0 | 0.4 | 0 | **0.6** |
| Schubert | Fantasie D760 1mov ''Wanderer''_38-98sec | 0 | **0.6** | 0.4 | 0 | 0 | **1** | 0 | 0 |
| | Fantasie D760 2mov ''Wanderer''_38-98sec | **0.6** | 0 | 0 | 0.4 | 0 | **0.6** | 0.4 | 0 |
| | Fantasie D760 3mov ''Wanderer''_38-98sec | 0 | **0.7** | 0 | 0.3 | 0 | **0.8** | 0 | 0.2 |
| | Moments Musicaux Op94 D780 N1_38-98sec | 0.2 | **0.8** | 0 | 0 | 0 | 0.2 | 0 | **0.8** |
| | Moments Musicaux Op94 D780 N2_38-98sec | 0 | 0 | 0.3 | **0.7** | 0 | **1** | 0 | 0 |

AB found the test hard as the MIDI recordings had taken out the "touché" (the piano player's keystroke dynamics) that is distinct between different composers and she had to rely purely on music pitch and rhythm for classification. The participant found Bach to be the easiest category due to its polyphonic structure and baroque harmonies. There were two cases of duck-rabbit scores, with Mozart's *Adagio in B flat* and *Another piece for piano* being rated as equally likely to be Beethoven/Schubert and Mozart/Beethoven respectively.

The performance of the CogAct model of AB is summarised in Table 2. AB had 8 correct categorisations on the 20 tests while the model had 14 correct categorisations. Focusing on the model, the Beethoven category had the highest mean confidence score of 0.66 and the highest proportion of correct answers (5/5). The corresponding mean confidence scores for the other composers were 0.40 for Mozart, 0.35 for Schubert and 0.54 for Bach. The proportion of correct categorisations was 3/5 for Mozart, 5/5 for Bach and 1/5 for Schubert. The Beethoven concept was the most activated one of the four composers – this was true for both the CogAct model (18/20) and AB (13/20). The least active concepts were Schubert for the CogAct model (4/20) and Bach for AB (6/20)

AB's prediction pairs were directly compared to the those of the model – summarised in Table 3.

*Table 3. The comparison metrics of AB and the CogAct model. Please see the Design of Simulations section for an explanation of the labels.*

| Composer | Excerpt | Identical | Both match | Tops match | One matches top | Single match |
|---|---|---|---|---|---|---|
| Bach | Bwv0525 Sonate En Trio N1_39-100sec | 0 | 0 | 1 | 1 | 1 |
| | Bwv0526 Sonate En Trio N2_39-100sec | 0 | 0 | 1 | 1 | 1 |
| | Bwv0527 Sonate En Trio N3_39-100sec | 0 | 0 | 1 | 1 | 1 |
| | Bwv0528 Sonate En Trio N4_39-100sec | 1 | 1 | 1 | 1 | 1 |
| | Bwv0529 Sonate En Trio N5_39-100sec | 0 | 0 | 1 | 1 | 1 |
| Beethoven | Anh06 Rondo_39-100sec | 0 | 0 | 0 | 0 | 1 |
| | Anh08Nb1 Gavotte 4 Hands_39-100sec | 0 | 1 | 0 | 1 | 1 |
| | Bagatella Op33 N1_39-100sec | 0 | 0 | 0 | 1 | 1 |
| | Bagatella Op33 N2_39-100sec | 0 | 0 | 0 | 0 | 1 |
| | Bagatella Op33 N3_39-100sec | 0 | 0 | 0 | 0 | 1 |
| Mozart | Adagio In B Flat_38-98sec | 0 | 0 | 0 | 0 | 1 |
| | Another Piece For Piano,K.176_38-98sec | 1 | 1 | 1 | 1 | 1 |
| | K19d Piano Sonata Duet_38-98sec | 1 | 1 | 1 | 1 | 1 |
| | K401 Fuga In G Minor _38-98sec | 0 | 0 | 0 | 1 | 1 |
| | K497 Piano Sonata 4 Hands_38-98sec | 0 | 0 | 0 | 0 | 1 |
| Schubert | Fantasie D760 1mov "Wanderer"_38-98sec | 0 | 0 | 1 | 1 | 1 |
| | Fantasie D760 2mov "Wanderer"_38-98sec | 0 | 0 | 0 | 0 | 0 |
| | Fantasie D760 3mov "Wanderer"_38-98sec | 1 | 1 | 1 | 1 | 1 |
| | Moments Musicaux Op94 D780 N1_38-98sec | 0 | 0 | 0 | 0 | 1 |
| | Moments Musicaux Op94 D780 N2_38-98sec | 0 | 0 | 0 | 0 | 0 |

Summing up, the comparison metrics were the following: 4/20 human and model prediction pairs were identical; 4/20 pairs matched disregarding the order of predictions; 9/20 of the top predictions matched; 12/20 model's prediction pairs contained the top human prediction; 18/20 model's prediction pairs contained at least one composer that was also predicted by the human.

### *Participant BC.*

Composer-specific piano music history (items with a star indicate answers where the participant was less than certain): Bach – the entire *Well Tempered Clavier (WTC)*, the entire *Goldberg Variations*; Beethoven – all 5 *Piano Concertos* and all 32 *Piano Sonatas* bar *Sonata No27*; Mozart – all *Piano Concertos* bar *K414*, all *Piano Sonatas\**, *Fantasia in C*; Schubert – all of the *Impromptus*, all *Piano Sonatas, Erlkonig D328, Fantasie D760*.

BC's performance and her CogAct model of is summarised in Table 4.

*Table 4. Categorisation of 21 novel music pieces by BC and her corresponding CogAct model. Numbers in bold signify the highest confidence score on a given test.*

| | File | CogAct model | | | | Human | | | |
|---|---|---|---|---|---|---|---|---|---|
| | | Bach | Beethoven | Mozart | Schubert | Bach | Beethoven | Mozart | Schubert |
| **Bach** | Bwv0525 Sonate En Trio N1_39-100sec | 0.4 | **0.6** | 0 | 0 | **1** | 0 | 0 | 0 |
| | Bwv0526 Sonate En Trio N2_39-100sec | 0.5 | **0.5** | 0 | 0 | **0.9** | 0 | 0.1 | 0 |
| | Bwv0527 Sonate En Trio N3_39-100sec | 0.5 | **0.5** | 0 | 0 | **1** | 0 | 0 | 0 |
| | Bwv0528 Sonate En Trio N4_39-100sec | 0.5 | **0.5** | 0 | 0 | **1** | 0 | 0 | 0 |
| | Bwv0529 Sonate En Trio N5_39-100sec | 0.4 | **0.6** | 0 | 0 | **1** | 0 | 0 | 0 |
| **Beethoven** | Bagatella Op33 N1_39-100sec | 0.2 | 0 | **0.8** | 0 | 0 | **0.9** | 0.1 | 0 |
| | Bagatella Op33 N2_39-100sec | 0.3 | **0.7** | 0 | 0 | 0 | **0.8** | 0 | 0.2 |
| | Bagatella Op33 N3_39-100sec | **0.5** | 0.5 | 0 | 0 | 0 | 0 | 0.2 | **0.8** |
| | Bagatella Op33 N4_39-100sec | 0 | 0.4 | **0.6** | 0 | 0 | 0 | **0.5** | 0.5 |
| | Bagatella Op33 N5_39-100sec | 0.3 | **0.7** | 0 | 0 | 0 | 0.2 | 0 | **0.8** |
| **Mozart** | Another Piece For Piano,K.176_38-98sec | 0.3 | 0 | **0.7** | 0 | 0 | 0 | **1** | 0 |
| | K354 Piano Variations "Je Suis Lindor"_38-98sec | 0.1 | 0 | **0.9** | 0 | 0 | 0.1 | **0.9** | 0 |
| | K398 Piano Variations "Salve Tu Domine"_38-98sec | 0.3 | 0 | **0.7** | 0 | 0 | **0.7** | 0.3 | 0 |
| | K401 Fuga In G Minor _38-98sec | 0.4 | **0.6** | 0 | 0 | **1** | 0 | 0 | 0 |
| | K573 9 Variations For Piano Uber Duport_38-98sec | 0.1 | 0 | **0.9** | 0 | 0 | 0 | **1** | 0 |
| **Schubert** | Moments Musicaux Op94 D780 N1_38-98sec | **0.5** | 0.5 | 0 | 0 | 0 | 0.2 | 0 | **0.8** |
| | Moments Musicaux Op94 D780 N2_38-98sec | 0 | 0 | 0.5 | 0.5 | 0 | 0.4 | 0 | **0.6** |
| | Moments Musicaux Op94 D780 N3_38-98sec | 0 | 0 | **0.8** | 0.2 | 0 | 0 | 0.4 | **0.6** |
| | Moments Musicaux Op94 D780 N4_38-98sec | 0.3 | 0 | **0.7** | 0 | **0.5** | 0 | 0.5 | 0 |
| | Moments Musicaux Op94 D780 N6_38-98sec | 0 | 0 | **0.7** | 0.3 | 0 | **0.7** | 0 | 0.3 |
| | Octet In Fmaj 3mov_38-98sec | 0 | 0 | **0.7** | 0.3 | 0 | 0.2 | 0 | **0.8** |

BC had 13 correct categorisations on the 21 tests and the model had 11 correct categorisations. Focusing on the model, the Mozart category had the highest mean confidence score of 0.64 and the highest proportion of correct answers (4/5). The corresponding mean confidence scores for the other composers were 0.46 for Beethoven, 0.22 for Schubert and 0.46 for Bach. The proportion of correct categorisations was 3/5 for Beethoven, 3/5 for Bach and 1/6 for Schubert.

BC's prediction pairs were directly compared to those of the model (see Table 5).

*Table 5. The comparison metrics of BC and the CogAct model.*

| | Excerpt | Identical | Both match | Tops match | One matches top | Single match |
|---|---|---|---|---|---|---|
| Bach | Bwv0525 Sonate En Trio N1_39-100sec | 0 | 0 | 0 | 1 | 1 |
| | Bwv0526 Sonate En Trio N2_39-100sec | 0 | 0 | 1 | 1 | 1 |
| | Bwv0527 Sonate En Trio N3_39-100sec | 0 | 0 | 1 | 1 | 1 |
| | Bwv0528 Sonate En Trio N4_39-100sec | 0 | 0 | 1 | 1 | 1 |
| | Bwv0529 Sonate En Trio N5_39-100sec | 0 | 0 | 0 | 1 | 1 |
| Beethoven | Bagatella Op33 N1_39-100sec | 0 | 0 | 0 | 0 | 1 |
| | Bagatella Op33 N2_39-100sec | 0 | 0 | 1 | 1 | 1 |
| | Bagatella Op33 N3_39-100sec | 0 | 0 | 0 | 0 | 0 |
| | Bagatella Op33 N4_39-100sec | 0 | 0 | 0 | 0 | 1 |
| | Bagatella Op33 N5_39-100sec | 0 | 0 | 0 | 0 | 1 |
| Mozart | Another Piece For Piano,K.176_38-98sec | 0 | 0 | 1 | 1 | 1 |
| | K354 Piano Variations ''Je Suis Lindor''_38-98sec | 0 | 0 | 1 | 1 | 1 |
| | K398 Piano Variations ''Salve Tu Domine''_38-98sec | 0 | 0 | 0 | 0 | 1 |
| | K401 Fuga In G Minor _38-98sec | 0 | 0 | 0 | 1 | 1 |
| | K573 9 Variations For Piano Uber Duport_38-98sec | 0 | 0 | 1 | 1 | 1 |
| Schubert | Moments Musicaux Op94 D780 N1_38-98sec | 0 | 0 | 0 | 0 | 1 |
| | Moments Musicaux Op94 D780 N2_38-98sec | 0 | 0 | 1 | 1 | 1 |
| | Moments Musicaux Op94 D780 N3_38-98sec | 0 | 1 | 0 | 1 | 1 |
| | Moments Musicaux Op94 D780 N4_38-98sec | 1 | 1 | 1 | 1 | 1 |
| | Moments Musicaux Op94 D780 N6_38-98sec | 0 | 0 | 0 | 0 | 1 |
| | Octet In Fmaj 3mov_38-98sec | 0 | 0 | 0 | 1 | 1 |

The comparison metrics were the following: 1/21 human and model prediction pairs were identical; 2/21 pairs matched disregarding the order of predictions; 9/21 of the top predictions matched; 14/21 model's prediction pairs contained the top human prediction; 20/21 model's prediction pairs contained at least one composer that was also predicted by the human.

### Participant CD.

Composer specific piano music history (items with a star indicate answers where the participant was less than certain): Bach – the entire *Well Tempered Clavier Volume 1 (WTC1)*, 20%* of *Well Tempered Clavier Volume 2 (WTC2),* the entire *Goldberg Variations, the entire English and French Suites*; Beethoven – all 5 *Piano Concertos* and most *Piano Sonatas\**, all *Bagatelles*; Mozart – most *Piano Concertos* except the early ones\*, all *Piano Sonatas\**, *Fantasia in C, Fantasia in D*, *Adagio in B flat*; Schubert – all of the *Impromptus*, some *Piano Sonatas\*,* all *Lieder*, some *Songs\*.*

CD's performance and her CogAct model is summarised in Table 6.

*Table 6. Categorisation of 21 novel music pieces by CD and her corresponding CogAct model. Numbers in bold signify the highest confidence score on a given test.*

| | | CogAct model | | | | Human | | | |
|---|---|---|---|---|---|---|---|---|---|
| | File | Bach | Beethoven | Mozart | Schubert | Bach | Beethoven | Mozart | Schubert |
| **Bach** | Bwv0525 Sonate En Trio N1_39-100sec | 0 | **0.6** | 0.4 | 0 | **1** | 0 | 0 | 0 |
| | Bwv0526 Sonate En Trio N2_39-100sec | **0.6** | 0.4 | 0 | 0 | **1** | 0 | 0 | 0 |
| | Bwv0527 Sonate En Trio N3_39-100sec | **0.6** | 0.4 | 0 | 0 | **1** | 0 | 0 | 0 |
| | Bwv0528 Sonate En Trio N4_39-100sec | **0.5** | 0 | 0.5 | 0 | **1** | 0 | 0 | 0 |
| | Bwv0529 Sonate En Trio N5_39-100sec | 0 | **0.6** | 0.4 | 0 | **1** | 0 | 0 | 0 |
| **Beethoven** | Anh06 Rondo_39-100sec | 0 | 0 | **0.6** | 0.4 | 0 | **0.6** | 0.4 | 0 |
| | Anh08Nb1 Gavotte 4 Hands_39-100sec | 0 | 0 | **0.8** | 0.2 | 0 | 0.2 | 0 | **0.8** |
| | Op077 Fantaisie_39-100sec | 0 | 0 | **0.5** | 0.5 | 0 | **0.7** | 0 | 0.3 |
| | Sonatina Fa Maj Woo_39-100sec | 0 | 0.4 | **0.6** | 0 | 0 | **0.7** | 0.3 | 0 |
| **Mozart** | Another Piece For Piano,K.176_38-98sec | 0.4 | 0 | **0.6** | 0 | 0 | 0 | **1** | 0 |
| | K19d Piano Sonata Duet_38-98sec | 0 | 0 | **0.8** | 0.2 | 0 | **1** | 0 | 0 |
| | K354 Piano Variations ''Je Suis Lindor''_38-98sec | 0.3 | 0 | **0.7** | 0 | 0 | 0 | **1** | 0 |
| | K398 Piano Variations ''Salve Tu Domine''_38-98sec | 0.3 | 0 | **0.7** | 0 | 0 | **1** | 0 | 0 |
| | K401 Fuga In G Minor _38-98sec | 0 | **0.7** | 0.3 | 0 | **1** | 0 | 0 | 0 |
| | K573 9 Variations For Piano Uber Duport_38-98sec | 0.3 | 0 | **0.7** | 0 | 0 | 0.2 | **0.8** | 0 |
| **Schubert** | Fantasie D760 1mov ''Wanderer''_38-98sec | 0 | **0.5** | 0.5 | 0 | **1** | 0 | 0 | 0 |
| | Fantasie D760 2mov ''Wanderer''_38-98sec | 0 | 0.4 | 0 | **0.6** | 0 | **0.6** | 0 | 0.4 |
| | Moments Musicaux Op94 D780 N1_38-98sec | 0.3 | **0.7** | 0 | 0 | 0 | 0 | 0 | **1** |
| | Moments Musicaux Op94 D780 N2_38-98sec | 0 | 0 | 0.4 | **0.6** | 0 | 0 | 0 | **1** |
| | Moments Musicaux Op94 D780 N4_38-98sec | 0.4 | 0 | **0.6** | 0 | 0 | 0 | 0 | **1** |
| | Moments Musicaux Op94 D780 N5_38-98sec | 0 | 0 | 0.4 | **0.6** | 0 | 0 | 0 | **1** |

CD had 15 correct categorisations on the 21 tests and the model had 11 correct categorisations. Focusing on the model, the Mozart category had the highest mean confidence score of 0.63 and the highest proportion of correct answers (5/6). The corresponding mean confidence scores for the other composers were 0.1 for Beethoven, 0.30 for Schubert and 0.34 for Bach. The proportion of correct categorisations was 0/4 for Beethoven, 3/5 for Bach and 3/6 for Schubert.

CD's prediction pairs were directly compared to the those of the model – summarised in Table 7.

*Table 7. The comparison metrics of CD and the CogAct model.*

| | Excerpt | Identical | Both match | Tops match | One matches top | Single match |
|---|---|---|---|---|---|---|
| Bach | Bwv0525 Sonate En Trio N1_39-100sec | 0 | 0 | 0 | 0 | 0 |
| | Bwv0526 Sonate En Trio N2_39-100sec | 0 | 0 | 1 | 1 | 1 |
| | Bwv0527 Sonate En Trio N3_39-100sec | 0 | 0 | 1 | 1 | 1 |
| | Bwv0528 Sonate En Trio N4_39-100sec | 0 | 0 | 1 | 1 | 1 |
| | Bwv0529 Sonate En Trio N5_39-100sec | 0 | 0 | 0 | 0 | 0 |
| Beethoven | Anh06 Rondo_39-100sec | 0 | 0 | 0 | 0 | 1 |
| | Anh08Nb1 Gavotte 4 Hands_39-100sec | 0 | 0 | 0 | 1 | 1 |
| | Op077 Fantaisie_39-100sec | 0 | 0 | 0 | 0 | 1 |
| | Sonatina Fa Maj Woo_39-100sec | 0 | 1 | 0 | 1 | 1 |
| Mozart | Another Piece For Piano,K.176_38-98sec | 0 | 0 | 1 | 1 | 1 |
| | K19d Piano Sonata Duet_38-98sec | 0 | 0 | 0 | 0 | 0 |
| | K354 Piano Variations "Je Suis Lindor"_38-98sec | 0 | 0 | 1 | 1 | 1 |
| | K398 Piano Variations "Salve Tu Domine"_38-98sec | 0 | 0 | 0 | 0 | 0 |
| | K401 Fuga In G Minor _38-98sec | 0 | 0 | 0 | 0 | 0 |
| | K573 9 Variations For Piano Uber Duport_38-98sec | 0 | 0 | 1 | 1 | 1 |
| Schubert | Fantasie D760 1mov "Wanderer"_38-98sec | 0 | 0 | 0 | 0 | 0 |
| | Fantasie D760 2mov "Wanderer"_38-98sec | 0 | 1 | 0 | 1 | 1 |
| | Moments Musicaux Op94 D780 N1_38-98sec | 0 | 0 | 0 | 0 | 0 |
| | Moments Musicaux Op94 D780 N2_38-98sec | 0 | 0 | 1 | 1 | 1 |
| | Moments Musicaux Op94 D780 N4_38-98sec | 0 | 0 | 0 | 0 | 0 |
| | Moments Musicaux Op94 D780 N5_38-98sec | 0 | 0 | 1 | 1 | 1 |

The comparison metrics were the following: 0/21 human and model prediction pairs were identical; 2/21 pairs matched disregarding the order of predictions; 8/21 of the top predictions matched; 11/21 model's prediction pairs contained the top human prediction; 13/21 model's prediction pairs contained at least one composer that was also predicted by the human.

**Participant DE.**

Composer specific piano music history (items with a star indicate answers where the participant was less than certain): Bach – the entire *Well Tempered Clavier Volume 1 (WTC1)*, all of *Well Tempered Clavier Volume 2 (WTC2)* bar *Preludes and Fugues* No 3, 9, 11,13, 17, 18; Beethoven – all *Piano Sonatas*; Mozart – half of *Piano Sonatas*\*; Haydn – about 10% of *Sonatas*\*.

DE's performance and her CogAct model is summarised in Table 8.

*Table 8. Categorisation of 20 novel music pieces by DE and her corresponding CogAct model. Numbers in bold signify the highest confidence score on a given test.*

| | File | CogAct model | | | | Human | | | |
|---|---|---|---|---|---|---|---|---|---|
| | | Bach | Beethoven | Haydn | Mozart | Bach | Beethoven | Haydn | Mozart |
| Bach | Bwv0525 Sonate En Trio N1_39-100sec | 0.3 | **0.7** | 0 | 0 | **0.7** | 0.15 | 0.15 | 0 |
| | Bwv0526 Sonate En Trio N2_39-100sec | **0.5** | **0.5** | 0 | 0 | 0.1 | **0.9** | 0 | 0 |
| | Bwv0527 Sonate En Trio N3_39-100sec | **0.6** | 0.4 | 0 | 0 | **0.7** | 0 | 0.15 | 0.15 |
| | Bwv0528 Sonate En Trio N4_39-100sec | **0.6** | 0.4 | 0 | 0 | **0.6** | 0.4 | 0 | 0 |
| | Bwv0529 Sonate En Trio N5_39-100sec | 0 | **0.7** | 0 | 0.3 | **0.8** | 0.2 | 0 | 0 |
| Beethoven | Bagatella Op33 N1_39-100sec | 0 | 0 | 0.4 | **0.6** | 0 | **0.6** | 0.4 | 0 |
| | Bagatella Op33 N2_39-100sec | 0.4 | **0.6** | 0 | 0 | 0 | **0.9** | 0 | 0.1 |
| | Bagatella Op33 N3_39-100sec | 0.2 | **0.8** | 0 | 0 | 0 | **0.5** | 0.5 | 0 |
| | Bagatella Op33 N4_39-100sec | 0 | 0 | 0.4 | **0.6** | 0 | **0.6** | 0.4 | 0 |
| | Bagatella Op33 N5_39-100sec | 0 | **0.9** | 0.1 | 0 | 0 | **0.8** | 0.2 | 0 |
| Haydn | 12 Menuets HobIX3_38-98sec | 0 | 0 | **0.5** | **0.5** | 0 | 0 | **0.7** | 0.3 |
| | Piano Piece N1_38-98sec | 0 | 0 | 0.3 | **0.7** | 0 | 0 | **0.7** | 0.3 |
| | Piano Piece N2_38-98sec | 0 | 0 | 0.4 | **0.6** | 0 | 0 | 0.4 | **0.6** |
| | Piano Piece N3_38-98sec | 0 | 0 | **0.7** | 0.3 | 0 | 0.3 | **0.7** | 0 |
| | Piano Piece N4_38-98sec | 0 | 0 | 0.2 | **0.8** | 0 | 0 | **0.6** | 0.4 |
| Mozart | Another Piece For Piano,K.176_38-98sec | 0 | 0 | 0.4 | **0.6** | 0 | 0.4 | **0.6** | 0 |
| | K354 Piano Variations ''Je Suis Lindor''_38-98sec | 0 | 0 | 0.3 | **0.7** | 0 | 0.33 | 0.33 | 0.33 |
| | K398 Piano Variations ''Salve Tu Domine''_38-98sec | 0 | 0 | 0.4 | **0.6** | 0 | **0.7** | 0.3 | 0 |
| | K401 Fuga In G Minor _38-98sec | 0.2 | **0.8** | 0 | 0 | **0.8** | 0.2 | 0 | 0 |
| | K573 9 Variations For Piano Uber Duport_38-98sec | 0 | 0 | 0.3 | **0.7** | 0 | 0 | 0.2 | **0.8** |

DE had 16 correct categorisations on the 20 tests and the model had 12 correct categorisations. DE was the only participant who completely abstained from using the "10/10" score, stating that "nothing in the world is 100%". Focusing on the model, the Mozart category had the highest mean confidence score of 0.52. The corresponding mean confidence scores for the other composers were 0.46 for Beethoven, 0.42 for Haydn and 0.40 for Bach. The proportion of correct categorisations was 3/5 for Beethoven, 3/5 for Bach and 2/5 for Haydn and 4/5 for Mozart. Bach was the least activated concept for both the model (7/20) and the human (6/20)

DE's participant prediction pairs were directly compared to the top two predictions of the model – summarised in Table 9.

*Table 9. The comparison metrics of DE and the CogAct model.*

|  | Excerpt | Identical | Both match | Tops match | One matches top | Single match |
|---|---|---|---|---|---|---|
| **Bach** | Bwv0525 Sonate En Trio N1_39-100sec | 0 | 0 | 0 | 1 | 1 |
|  | Bwv0526 Sonate En Trio N2_39-100sec | 0 | 1 | 0 | 1 | 1 |
|  | Bwv0527 Sonate En Trio N3_39-100sec | 0 | 0 | 1 | 1 | 1 |
|  | Bwv0528 Sonate En Trio N4_39-100sec | 1 | 1 | 1 | 1 | 1 |
|  | Bwv0529 Sonate En Trio N5_39-100sec | 0 | 0 | 0 | 0 | 1 |
| **Beethoven** | Bagatella Op33 N1_39-100sec | 0 | 0 | 0 | 0 | 1 |
|  | Bagatella Op33 N2_39-100sec | 0 | 0 | 1 | 1 | 1 |
|  | Bagatella Op33 N3_39-100sec | 0 | 0 | 0 | 0 | 1 |
|  | Bagatella Op33 N4_39-100sec | 0 | 0 | 0 | 0 | 1 |
|  | Bagatella Op33 N5_39-100sec | 1 | 1 | 1 | 1 | 1 |
| **Haydn** | 12 Menuets HobIX3_38-98sec | 0 | 1 | 0 | 1 | 1 |
|  | Piano Piece N1_38-98sec | 0 | 1 | 0 | 1 | 1 |
|  | Piano Piece N2_38-98sec | 1 | 1 | 1 | 1 | 1 |
|  | Piano Piece N3_38-98sec | 0 | 0 | 1 | 1 | 1 |
|  | Piano Piece N4_38-98sec | 0 | 1 | 0 | 1 | 1 |
| **Mozart** | Another Piece For Piano,K.176_38-98sec | 0 | 0 | 0 | 1 | 1 |
|  | K354 Piano Variations "Je Suis Lindor"_38-98sec | 1 | 1 | 1 | 1 | 1 |
|  | K398 Piano Variations "Salve Tu Domine"_38-98sec | 0 | 0 | 0 | 0 | 1 |
|  | K401 Fuga In G Minor _38-98sec | 0 | 1 | 0 | 1 | 1 |
|  | K573 9 Variations For Piano Uber Duport_38-98sec | 1 | 1 | 1 | 1 | 1 |

The comparison metrics were the following: 5/20 human and model prediction pairs were identical; 10/20 pairs matched disregarding the order of the predictions; 8/20 of the top predictions matched; 15/20 model's prediction pairs contained the top human prediction; 20/20 model's prediction pairs contained at least one composer that was also predicted by the human.

**Participant EF.**

Composer-specific piano music history (items with a star indicate answers where the participant was less than certain): Bach – the entire *Well Tempered Clavier (WTC)*, the entire *Goldberg Variations*, all *Piano Concertos*, all *Piano Suites*; Beethoven – all *Piano Sonatas\**; Mozart – most *Piano Concertos\**, all *Piano Sonatas\**, *Fantasia in C*, *Fantasia in D*; Schubert – all of the *Impromptus*, only 3 *Piano Sonatas\**.

EF's performance and her CogAct model is summarised in Table 10.

*Table 10. Categorisation of 20 novel music pieces by EF and her corresponding CogAct model. Numbers in bold signify the highest confidence score on a given test.*

| | File | CogAct model | | | | Human | | | |
|---|---|---|---|---|---|---|---|---|---|
| | | Bach | Beethoven | Mozart | Schubert | Bach | Beethoven | Mozart | Schubert |
| Bach | Bwv0527 Sonate En Trio N3_39-100sec | 0 | 0.4 | **0.6** | 0 | **1** | 0 | 0 | 0 |
| | Bwv0528 Sonate En Trio N4_100_112m | 0.5 | 0.5 | 0 | 0 | **1** | 0 | 0 | 0 |
| | Bwv0529 Sonate En Trio N5_194_210m | **0.6** | 0.4 | 0 | 0 | **1** | 0 | 0 | 0 |
| | Bwv0963 Sonata_39-100sec | 0 | 0.5 | 0 | 0.5 | **1** | 0 | 0 | 0 |
| | Bwv0967 Sonata_39-100sec | **0.6** | 0.4 | 0 | 0 | **1** | 0 | 0 | 0 |
| Beethoven | Anh06 Rondo_79_143m | 0 | **0.7** | 0.3 | 0 | 0 | **0.7** | 0.3 | 0 |
| | Sonatina In C_103_166m | 0.4 | **0.6** | 0 | 0 | 0 | **0.9** | 0.1 | 0 |
| | Sonatina Op33 4mov_38_105m | 0.3 | **0.7** | 0 | 0 | 0 | 0.5 | 0.5 | 0 |
| | Sonatina Op79_39-100sec | 0 | 0.4 | **0.6** | 0 | 0 | **1** | 0 | 0 |
| | Sonatina WoO050_39-100sec | 0 | 0.4 | **0.6** | 0 | 0 | 0 | **1** | 0 |
| Mozart | Adagio In B Flat_38-98sec | 0 | 0.2 | **0.8** | 0 | 0 | 0.1 | **0.9** | 0 |
| | K398 Piano Variations ''Salve Tu Domine''_38-98sec | 0 | 0.3 | **0.7** | 0 | 0 | 0.3 | **0.7** | 0 |
| | K573 9 Variations For Piano Uber Duport_110_170m | 0.5 | 0.5 | 0 | 0 | 0 | **0.8** | 0.2 | 0 |
| | Sonatina N21 3mov_7_53m | 0.4 | **0.6** | 0 | 0 | 0 | 0.5 | 0.5 | 0 |
| | Sonatina N22 4mov_3_62m | 0.5 | 0.5 | 0 | 0 | 0 | 0.2 | **0.8** | 0 |
| Schubert | Moments Musicaux Op94 D780 N1_38-98sec | 0.4 | **0.6** | 0 | 0 | 0 | 0.2 | 0 | **0.8** |
| | Moments Musicaux Op94 D780 N2_38-98sec | 0 | 0 | 0.5 | 0.5 | 0 | 0 | 0 | **1** |
| | Moments Musicaux Op94 D780 N3_22_68m | 0 | 0.5 | 0.5 | 0 | 0 | 0.3 | 0 | **0.7** |
| | Moments Musicaux Op94 D780 N4_17_74m | 0.4 | **0.6** | 0 | 0 | 0 | **0.5** | 0 | 0.5 |
| | Moments Musicaux Op94 D780 N5_38-98sec | 0 | 0.3 | **0.7** | 0 | 0 | **0.5** | 0 | 0.5 |

EF had 18 correct categorisations on the 20 tests and the model had 9 correct categorisations. Focusing on the model, the Beethoven category had the highest mean confidence score of 0.56. The corresponding mean confidence scores for the other composers were 0.34 for Bach, 0.25 for Mozart and 0.10 for Schubert. The proportion of correct categorisations was 3/5 for Beethoven, 3/5 for Bach and 1/5 for Schubert and 2/5 for Mozart.

EF's prediction pairs were directly compared to the top two predictions of the model (see Table 11).

*Table 11. The comparison metrics of EF and the CogAct model.*

| | Excerpt | Identical | Both match | Tops match | One matches top | Single match |
|---|---|---|---|---|---|---|
| Bach | Bwv0527 Sonate En Trio N3_39-100sec | 0 | 0 | 0 | 0 | 0 |
| | Bwv0528 Sonate En Trio N4_100_112m | 0 | 0 | 1 | 1 | 1 |
| | Bwv0529 Sonate En Trio N5_194_210m | 0 | 0 | 1 | 1 | 1 |
| | Bwv0963 Sonata_39-100sec | 0 | 0 | 0 | 0 | 0 |
| | Bwv0967 Sonata_39-100sec | 0 | 0 | 1 | 1 | 1 |
| Beethoven | Anh06 Rondo_79_143m | 1 | 1 | 1 | 1 | 1 |
| | Sonatina In C_103_166m | 0 | 0 | 1 | 1 | 1 |
| | Sonatina Op33 4mov_38_105m | 0 | 0 | 0 | 0 | 1 |
| | Sonatina Op79_39-100sec | 0 | 0 | 0 | 1 | 1 |
| | Sonatina WoO050_39-100sec | 0 | 0 | 1 | 1 | 1 |
| Mozart | Adagio In B Flat_38-98sec | 1 | 1 | 1 | 1 | 1 |
| | K398 Piano Variations "Salve Tu Domine"_38-98sec | 1 | 1 | 1 | 1 | 1 |
| | K573 9 Variations For Piano Uber Duport_110_170m | 0 | 0 | 0 | 1 | 1 |
| | Sonatina N21 3mov_7_53m | 0 | 0 | 0 | 0 | 1 |
| | Sonatina N22 4mov_3_62m | 0 | 0 | 0 | 0 | 1 |
| Schubert | Moments Musicaux Op94 D780 N1_38-98sec | 0 | 0 | 0 | 0 | 1 |
| | Moments Musicaux Op94 D780 N2_38-98sec | 0 | 0 | 1 | 1 | 1 |
| | Moments Musicaux Op94 D780 N3_22_68m | 0 | 0 | 0 | 0 | 1 |
| | Moments Musicaux Op94 D780 N4_17_74m | 0 | 0 | 0 | 0 | 1 |
| | Moments Musicaux Op94 D780 N5_38-98sec | 0 | 0 | 0 | 0 | 1 |

The comparison metrics were the following: 3/20 human and model prediction pairs were identical; 3/20 pairs matched disregarding the order of the predictions; 9/20 of the top predictions matched; 11/20 model's prediction pairs contained the top human prediction; 18/20 model's prediction pairs contained at least one composer that was also predicted by the human.

### Participant FG.

Composer specific piano music history (items with a star indicate answers where the participant was less than certain): Beethoven – all *Piano Sonatas* bar *No9*, *No15*, *No16*, *No19*, *No20*, *No22*, *No24*, *No25*; Mozart – all *Piano Concertos*, *Piano Sonatas No5*, *No 6*, *No8*, *No11*, *No13-16*, *K381 for four hands*, *Fantasia in C*, *Fantasia in D*; Haydn – *Concerto in D minor*, 35% of *Sonatas\**; Schubert – all of the *Impromptus*, all of the *Klavierstucke* collection, *Piano Sonatas No14, No16, No20, No21*.

FG's performance and her CogAct model is summarised in Table 12.

*Table 12. Categorisation of 21 novel music pieces by FG and her corresponding CogAct model. Numbers in bold signify the highest confidence score on a given test.*

| | File | CogAct model | | | | Human | | | |
|---|---|---|---|---|---|---|---|---|---|
| | | Beethoven | Haydn | Mozart | Schubert | Beethoven | Haydn | Mozart | Schubert |
| Beethoven | Piano Sonata N09_78_141m | 0.8 | 0 | 0.2 | 0 | 0 | 0.7 | 0.3 | 0 |
| | Piano Sonata N15_71_163m | 0.7 | 0.3 | 0 | 0 | 0.8 | 0.2 | 0 | 0 |
| | Piano Sonata N16_53_158M | 0.8 | 0 | 0.2 | 0 | 0 | 0.2 | 0.8 | 0 |
| | Piano Sonata N19_14_69m | 0.9 | 0 | 0.1 | 0 | 0.8 | 0 | 0.2 | 0 |
| | Piano Sonata N20_14_72m | 0.5 | 0.5 | 0 | 0 | 0.7 | 0 | 0 | 0.3 |
| Haydn | 12 Menuets HobIX3_9_67m | 0.8 | 0.2 | 0 | 0 | 0 | 0.2 | 0.8 | 0 |
| | 20 Variations For Keyboard HobXVII2_174_226m | 0 | 0.6 | 0.4 | 0 | 0 | 0.3 | 0.7 | 0 |
| | Capriccio_2_83m | 0 | 0.5 | 0.5 | 0 | 0.3 | 0.7 | 0 | 0 |
| | Haydn_43_1_8_64m | 0 | 0.7 | 0.3 | 0 | 0.6 | 0.4 | 0 | 0 |
| | Haydn_Sonata_No43_2mov_27_89m | 0 | 0.7 | 0.3 | 0 | 0 | 0.2 | 0.8 | 0 |
| Mozart | Adagio In B Flat_2_17m | 0 | 0.5 | 0.5 | 0 | 0.4 | 0.6 | 0 | 0 |
| | K398 Piano Variations ''Salve Tu Domine''_20_61 | 0 | 0.6 | 0.4 | 0 | 1 | 0 | 0 | 0 |
| | K401 Fuga In G Minor_7_63m | 0.8 | 0.2 | 0 | 0 | 0.5 | 0.5 | 0 | 0 |
| | K573 9 Variations For Piano Uber Duport_10_78m | 0 | 0.5 | 0.5 | 0 | 0 | 0 | 1 | 0 |
| | Piano Sonata N02 K280_55_126m | 0 | 0.6 | 0.4 | 0 | 0.6 | 0.4 | 0 | 0 |
| | Piano Sonata N03 K281_9_59m | 0.4 | 0.6 | 0 | 0 | 1 | 0 | 0 | 0 |
| Schubert | Fantasie D760 1mov ''Wanderer''_3_68M | 0.9 | 0 | 0 | 0.1 | 0.7 | 0.3 | 0 | 0 |
| | Fantasie D760 2mov ''Wanderer''_1_13m | 0 | 0.5 | 0 | 0.5 | 0.4 | 0 | 0 | 0.6 |
| | Piano Rondo Op107 D951_8_41m | 0.4 | 0.6 | 0 | 0 | 0.8 | 0.2 | 0 | 0 |
| | Piano Sonata D568 Op.Posth.122_12_82m | 0 | 0.5 | 0.5 | 0 | 1 | 0 | 0 | 0 |
| | Piano Sonata N04 D537 Op164_25_101 | 0 | 0.7 | 0.3 | 0 | 0 | 0 | 0 | 1 |

FG had 7 correct categorisations on the 21 tests and the model had 12 correct categorisations. Focusing on the model, the Beethoven category had the highest mean confidence score of 0.74. The corresponding mean confidence scores for the other composers were 0.54 for Haydn, 0.30 for Mozart and 0.12 for Schubert. The proportion of correct categorisations was 5/5 for Beethoven, 4/5 for Haydn and 1/5 for Schubert and 2/6 for Mozart. Schubert was the least activated concept for both the model (2/21) and the human (3/21).

FG's prediction pairs were directly compared to the predictions of the model – summarised in Table 13.

*Table 13. The comparison metrics of FG and the CogAct model.*

| | Excerpt | Identical | Both match | Tops match | One matches top | Single match |
|---|---|---|---|---|---|---|
| Beethoven | Piano Sonata N09_78_141m | 0 | 0 | 0 | 0 | 1 |
| | Piano Sonata N15_71_163m | 1 | 1 | 1 | 1 | 1 |
| | Piano Sonata N16_53_158M | 0 | 0 | 0 | 1 | 1 |
| | Piano Sonata N19_14_69m | 1 | 1 | 1 | 1 | 1 |
| | Piano Sonata N20_14_72m | 0 | 0 | 0 | 1 | 1 |
| Haydn | 12 Menuets HobIX3_9_67m | 0 | 0 | 0 | 0 | 1 |
| | 20 Variations For Keyboard HobXVII2_174_226m | 0 | 1 | 0 | 1 | 1 |
| | Capriccio_2_83m | 0 | 0 | 1 | 1 | 1 |
| | Haydn_43_1_8_64m | 0 | 0 | 0 | 0 | 1 |
| | Haydn_Sonata_No43_2mov_27_89m | 0 | 1 | 0 | 1 | 1 |
| Mozart | Adagio In B Flat_2_17m | 0 | 0 | 1 | 1 | 1 |
| | K398 Piano Variations "Salve Tu Domine"_20_61 | 0 | 0 | 0 | 0 | 1 |
| | K401 Fuga In G Minor_7_63m | 0 | 1 | 0 | 1 | 1 |
| | K573 9 Variations For Piano Uber Duport_10_78m | 0 | 0 | 0 | 1 | 1 |
| | Piano Sonata N02 K280_55_126m | 0 | 0 | 0 | 0 | 1 |
| | Piano Sonata N03 K281_9_59m | 0 | 0 | 0 | 1 | 1 |
| Schubert | Fantasie D760 1mov "Wanderer"_3_68M | 0 | 0 | 1 | 1 | 1 |
| | Fantasie D760 2mov "Wanderer"_1_13m | 0 | 0 | 0 | 1 | 1 |
| | Piano Rondo Op107 D951_8_41m | 0 | 1 | 0 | 1 | 1 |
| | Piano Sonata D568 Op.Posth.122_12_82m | 0 | 0 | 0 | 0 | 0 |
| | Piano Sonata N04 D537 Op164_25_101 | 0 | 0 | 0 | 0 | 0 |

The comparison metrics were the following: 2/21 human and model prediction pairs were identical; 6/21 pairs matched disregarding the order of the predictions; 5/21 of the top predictions matched; 14/21 model's prediction pairs contained the top human prediction; 18/21 model's prediction pairs contained at least one composer that was also predicted by the human.

### *Statistical analysis of the combined CogAct models.*

The detailed calculation of statistical significance of the combined CogAct models is presented can be found in the supplementary online material section at https://github.com/Voskod/CogAct.

To generate the cumulative/total metric scores, individual metric scores were summed across all the participants. The resulting cumulative/total metric scores *for*

*Identical, Both match, Tops match, One matches top* and *Single Match* were 15, 27, 48, 77 and 107 respectively.

The Bernoulli formula for discrete binomial distribution was used to estimate the statistical significance of the results. The p-values of the metrics were calculated as follows:

$$P(exactly\,k) = p^k * (1-p)^{n-k} * \binom{n}{k} \tag{1}$$

$$P(at\,least\,k) = \sum_{i=k}^{n} P(exactly\,i) \tag{2}$$

Here, *p* is the probability of the corresponding metric getting a match with the human score, *n* is the number of trials and *k* is the number of matching answers.

P(at least 15 "Identical" answers) = 0.00529

P(at least 27 "Both match" answers) < 0.001

P(at least 48 "Tops match" answers) < 0.001

P(at least 77 "One matches top" answers) < 0.001

P(at least 107 "Single match' answers) < 0.001

(See supplementary online material for the exact p-values that were less than 0.001.)

To adjust for multiple hypothesis testing, a Bonferroni adjustment was made. Since there were 5 hypotheses and the confidence threshold was 0.05, the adjusted confidence threshold became 0.05 / 5 = 0.01. Since the p-values above fall into the 0.01 threshold we may now reject the null hypothesis that the observed number occurred by chance and accept the experimental hypothesis: CogAct models of music concept formation produce a good fit to the human data.

## CogAct's correspondence to a deep learning model

The currently ongoing AI revolution is powered by formal models of artificial neural networks (ANNs, historically known under umbrella terms of connectionism and, more recently, deep learning)(Hambling, 2020; Jo, Nho, & Saykin, 2019; Mnih et al., 2013; Silver et al., 2016; Silver et al., 2017). Deep learning is commonly used in psychological models (e.g., Battleday et al., 2020; Hoffman, McClelland, & Lambon Ralph, 2018; Sanders & Nosofsky, 2020) and may also be a form of a UTC. Indeed, its set of fundamental mechanisms was largely developed and refined through research in neuroscience and psychology (Hahnloser et al., 2000; Hinton & McClelland, 1987; Hinton et al., 2012; McCulloch & Pitts, 1943; Nair & Hinton, 2010; Rosenblatt, 1958, 1962; Rumelhart, Hinton, & Williams, 1986). Moreover, deep learning models have a rich history in the field of categorisation. For example, over 40 text classification datasets and 150 deep learning models have been built in recent years alone (Minaee et al., 2020). Summatively, psychological plausibility and previous research into categorisation made deep learning an obvious choice for a baseline comparison to CogAct, above other alternative classification algorithms.

We should stress that our baseline deep learning model is meant to be supplementary to CogAct research – while our ANN model is not trivial, it makes no claim to be state-of-the-art. Instead, it forms a performance baseline and gives important theoretical neuroscience context to our state-of-the-art cognitive model of concept learning (CogAct). We should further note, that our ANN model cannot adapt to concept complexity and overfits/fails in simpler category learning tasks described in our previous section.

We chose a recurrent neural network (RNN) architecture for our ANN model. It contained six layers (Embedding, RNN, Flatten, Dropout, and two Dense layers) with 8,320,207 trainable parameters.

 A common problem with sequence learning RNNs is that the network "forgets" input that is above approximately ten time steps (Bengio, Frasconi, & Schmidhuber, 2001; Goodfellow, Bengio, & Courville, 2016). Thus, we developed an ANN attention with LTM activation mechanism similar to that of CogAct. The sliding attention window passes the retrieved short word sequences to the model; the model generates a vote/prediction for each of the sequences; these votes are then aggregated and the overall winner is declared by the confidence formula. The multidomain confidence criterion was calculated by the same formula as was used in CogAct model and still had the aim of resolving conflicting "voting" among category representations. The important difference was that, this time, the voting conflict was among different neural activations/engrams (as opposed to different cognitive chunks with CogAct):

$$C(c_i|x) = a_i \Big/ \sum_{k=1}^{m} (a_k)$$

where $C(c_i|x)$ is confidence that category label (i.e., a composer) is $c_i$ , given a novel music score $x$; $a_i$ is the neural engrams' activation score corresponding to that category (i.e., a composer), and the summation part being the sum of neural engram activations across all $m$ categories.

We trained the ANN via back propagation. The ANN model was tasked with the same categorisation task as CogAct, with each participant being represented by an individual bespoke ANN model. For further details and code see https://github.com/Voskod/CogAct.

**The fit of the ANN models to the human data.**

The "per participant" tables and the calculation of statistical significance of the combined ANN models is presented in the supplementary online material at https://github.com/Voskod/CogAct.

The resulting ANN cumulative metric scores *for Identical, Both match, Tops match, One matches top* and *Single Match* were 11, 19, 51, 78 and 110 respectively.

To adjust for multiple hypothesis testing, a Bonferroni adjustment was made. Since there were 5 hypotheses and the confidence threshold was 0.05, the adjusted confidence threshold became 0.05 / 5 = 0.01. Since the 3 of the 5 metrics' p-values fall into the 0.01 threshold, we may now reject the null hypothesis and accept the experimental hypothesis: ANN models of music concept formation produce a good fit to the human data.

## Discussion

The results of the study can be summarised as follows: CogAct models of music concept formation produced good fit to human categorisation performance. CogAct models performed better on the two most stringent comparison metrics (*Identical* and *Both match*) and had similar performance to our baseline ANN models on the other three metrics (*Tops match*, *One matches top* and *Single match*).

There are four key strengths and contributions of the current study.

Firstly, we produced a direct three-way comparison between human, cognitive psychology-based model (CogAct), and neuroscience-based model (ANN). These further bridges the gap between cognition and its neural substrate, mutually reinforces the positions of the chunking theory, connectionism, and modern probabilistic approach to concepts. The current methodology may be easily extended to experiments in other areas of psychology.

Secondly, we moved away from modelling "the average human" and accounted for individual differences in prior knowledge of each of its participants and their respective models. Thirdly, the models moved away from relying on hand engineered features/dimensions while learning from complex real-life data. The fourth strength of the study was the choice of the music modality as the conceptual domain in focus – the abstract nature of music vocabulary and semantics allowed to significantly diminish the confounding effects of reductionism inherent in modelling concept formation in language-based domains.

As was mentioned above, the choice of classical music stimuli in the current study allowed for rigorous control of the training data/history, while maintaining the ecological validity of the natural category stimuli. Of course, the increased rigour in controlling training data was not absolute due to the reliance on self-reported history. Indeed, this shortcoming was highlighted by case of the Participant CD. She did not report Schubert's *Moments Musicaux* and *Fantasie "Wanderer"* as part of her musical history during the first stage of the interviews, only to say that the test pieces (which *were* excerpts from *Moments Musicaux* and *Fantasie "Wanderer"*) reminded her of the pieces above. However, most participants have specifically stated that the test pieces were in fact novel to them.

There were two areas where the divergence between the models and the corresponding human participants was particularly notable. First, the number of instances of highly confident judgments were radically different between the human participants and the models. Approximately 1/3 (40/122) of all human ratings were 100% confidence scores. In contrast, CogAct models had no 100% ratings at all and ANNs had only 2 (out of 122). With that said, human participant DE explicitly stated that "nothing in the world is 100%" and abstained from 100% confidence ratings. Thus, the models' "hedging of bets" may not have been completely unrealistic, even though this aspect of subjective judgment was not captured.

 Secondly, human participants had almost no Bach concept activation across non-Bach stimuli. For both model types, however, Bach tended to be among the most active concepts throughout. This led to a substantial level of misses on the "Identical" and "Both match" prediction scores' metrics.

A subsequent study could retest the same human participants on the same test stimuli. Indeed, human performance is not a point, but a curve (Barsalou, 1987; Heitz, 2014). For example, in his study of concept stability, Barsalou (1989) found that the same participants produced only 66% of identical concept properties over two sessions separated by a two-week break. The stability of musical prediction pairs would thus similarly warrant further investigation.

Another potentially significant weakness of the models was the dismissed timing in music. One extension to the current study would be to incorporate rhythm into the models. With that said, the MIDI synthesiser playing the musical excerpts to the human participants had also distorted the rhythm of some of the pieces. The Participant CD reported the distortion to be particularly noticeable but agreed that the note/chord pitch values were much more important for categorisation than the rhythm.

The exclusive focus on piano music might also have been problematic. This was highlighted by the Participant FG who classified one novel piece as a Beethoven *because* it reminded her of a Beethoven Symphony. A follow-up study that includes non-piano pieces of the participants' musical history could be justified. More broadly, the modelled musical history may also include composers that lie outside of the test set.

Lastly, the current approach to modelling individual concept spaces was based purely on individual differences in musical history. An extension to the current study could incorporate other individual differences. For example, participants' STM span could be measured prior to categorisation experiment. The STM size of the model may then match that of the corresponding participant to achieve a closer fit of the model.

# General Discussion

In recent years, research on concept formation has been dominated by the prototype, exemplar, and clustering-based approaches. In spite of considerable empirical and theoretical advances, two related factors have hampered progress. First, outside of deep learning, most computational models in the field require that the features to learn are hand-coded and that exemplars are stored in a bespoke database (e.g., Braunlich & Love, 2022; Lieto, 2019; Love et al., 2004; Nosofsky, Sanders, & McDaniel, 2018; Sanders & Nosofsky, 2020). Put differently, unlike humans, models are not able to learn concepts from raw, ecological input. Second, theories have been developed to account for average data, not individual data. Thus, models have tended to simulate behaviour averaged across participants when categorising simple material. In this paper, we have adopted a converse approach: the simulation of individual behaviour when categorising complex material.

Our first aim was to develop a theory of concept formation that explains behaviour with a complex, raw and ecological input; specifically, features were not preset. Our extension of the chunking theory UTC that uses chunking as learning mechanism, was implemented in a computer program. It seamlessly integrates mechanisms related to perception, memory, and learning to account for data in concept-formation experiments.

Our second aim was to develop a novel methodology for evaluating categorisation models: models learn concepts using individualised input that matches as far as possible the learning material that was used by each human participant. Thus, the CogAct models were trained using individualised input, and tested with the same material as the human participants; this material differed between models as it was tailored to the material that each participant used during their own training and testing. This approach allowed us to simulate the subjective conceptual apparatus of each participant individually. It is important that both the human study and the simulations took part in in the same high-dimensional natural domain.

To validate the models, CogAct was used in a number of categorisation experiments, from learning simple artificial concepts to learning complex natural concepts from raw, naturalistic input. This addressed Battleday et al.'s (2020) key criticism of transferring conclusions about model performance developed for low-dimensional stimuli to higher-dimensional/natural categories. Offering a proof of sufficiency (Newell & Simon, 1972), the models successfully learnt artificial concepts in the binary logic task, the five-four task, and sequential data with occlusion (the latter had previously been a problem for chunking models). CogAct's ability to learn natural concepts from raw input was further established in three domains: literature, chess and music. Critically, with music, CogAct learnt concepts at a level similar to that of the human participants being modelled, without bootstrapping to human devised task-specific pair-wise judgments or other hand-made knowledge structures. In broader terms, this means that the all-important definition of a concept (e.g., a mental representation such as "XOR", "Mozart" or "Sicilian defence") may be reformulated as a collection of chunks in a cognitive LTM-like structure – as operationalised by chunking theory's CogAct.

## Range of domains explained by chunking

Chunking theory is among the most established theories in cognitive psychology. As was discussed above, cognitive chunking has been found to be central in numerous domains, ranging from verbal learning, to perception and memory of experts, to emotion processing, grammar acquisition, reasoning, cognitive decline due to ageing – and the list goes on. An important contribution of our research is to have extended the range of domains where chunking has been successfully applied, including adaptation to category structure and categorisation from complex raw input in dissimilar domains of music, chess and literature. This makes CogAct important both as a stand-alone model of concept learning and as a way of integrating concept learning into broader psychological mechanisms (often difficult for other concept formation models, e.g., based on exemplar, prototype, Bayesian or other types of clustering). Several characteristics made this extension possible. First, the same model was used in all simulations. It was able to pick up the conceptual structure of the domain from the stimuli, and was able to adapt to the complexity of the task automatically and seamlessly.

Second, CogAct integrates high-level perception, memory, and learning mechanisms. Importantly, as raw data were used, it was not necessary to pre-code features. Thus, there is no "concept formation" mechanism per se: the CogAct model for acquiring concepts essentially uses the same architecture as the model employed, e.g., for simulating expert memory. Thus, a key advantage of CogAct is that models learning with a similar chunking mechanism had already simulated a number of phenomena in fields such as verbal learning (Richman & Simon, 1989; Richman et al., 2002), the acquisition of language (vocabulary and syntactic categories) (Freudenthal et al., 2016; Gegov et al., 2012; Jones et al., 2010), and the acquisition of expertise (Gobet & Simon, 1998; Guida et al., 2012b).

Together, these results place chunking as a candidate mechanism for explaining concept formation, along with the explanations offered by prototype, exemplar, and clustering-based theories.

Third, the research established the general nature of the chunking learning mechanism, a fairly simple mechanism. Thus, considering its successful application to model different domains, from language acquisition to expertise to concept formation, CogAct is an existence proof that chunking is a serious candidate for a general learning mechanism, in line with the single algorithm hypothesis (Hawkins & Blakeslee, 2004; LeCun et al., 2015; Mountcastle, 1978) and unified theories of cognition (Byrne, 2012; Newell, 1990). As chunking theory also accounts for data in a number of domains, we suggest that chunking is a general and omnipresent mechanism, and a key candidate for a fundamental mechanism of human cognition.

## Similarity between chunking theory and deep learning

The comparative performances of the CogAct and ANN models provide for intriguing analysis and warrant further investigation. From the high-level birds-view perspective, both models demonstrated the capability of learning concepts in a complex music domain (although, unlike CogAct, our ANN model cannot adapt). Both CogAct and ANN models simulated human categorisation performance with broadly similar levels of goodness of fit to the human data.

Apart from qualitative and quantitative similarities in predictions, CogAct and ANNs also share functional similarities. This was demonstrated through the use of the *same* activation mechanism for both CogAct and ANN models. Indeed, the proposed activation formula/code was completely interchangeable between the models and required no adjustment to measure the activation of *chunks* or the activation of *neural engrams*. In this technical sense, cognitive chunks may be said to be *equivalent* to neural engrams.

This similarity between CogAct and deep learning supports the hypothesis of a single learning mechanism. In the previous section, this hypothesis was discussed with respect to a single model across several tasks. It is now discussed across different modelling traditions.

What would be the theoretical justification for using a single learning mechanism? We suggest that such an algorithm must essentially pick up the statistical structure of the environment and that, as noted for example by Simon (1969), understanding the structure of the environment goes a long way in predicting behaviour. In other words, our argument is that any algorithm strong enough would do the job (see also Marr's (1982) idea of computational level). Unless one focuses on more detailed data – e.g., reaction times and patterns of errors – the detail of the algorithm does not matter much, as long as it is strong enough to pick up the statistical regularities of the environment.

CogAct is technically a symbolic model. However, there are several dimensions along which it is actually similar to non-symbolic architectures such as artificial neural networks. These include the presence of a single learning mechanism and the ability to use raw data as input. Another important similarity is the shared universal function approximator property of deep learning (Funahashi, 1989; Haykin, 1999; Hornik, Stinchcombe, & White, 1989) and chunking networks (Fredkin, 1960; Gobet, 1996). The results of the current study also offer further support to the view that the divide between subsymbolic and symbolic architectures is not as clearcut as often assumed (e.g., Richman & Simon, 1989; Simon, 1991, pp. 81-82).

## Individualised models

An important methodological contribution of this article is to have shown the power of using individualised models, where both the training input and the material used during testing are matched to that used by human participants. Our methodology allowed for greater control over prior knowledge in simulations of human performance. While it is difficult or even impossible to determine a human's history with regard to her "training data" in many natural domains, formally studied music does allow for a relatively precise estimate of participants' experience. This approach voids the drawback of "modelling the average participant"; thus, rather than losing information because it is considered as noise, the models can explain at least some of the individual differences (Gobet, 2017; Gobet & Ritter, 2000; Newell, 1973).

In this article, we have shown how CogAct models can simulate subjective judgments by taking account of differences in prior knowledge. For example, CogAct model of Participant EF categorised Bach's *Bwv0526 Sonate en trio no5* as a Bach, but CogAct model of participant DE categorised this piece as a Beethoven. The overlap of the subjective judgments of the models was also evident – for example, in the Mozart's *Fugue in G* case, with both models AB and BC judging the above piece to be Beethoven's. Having said that, it is important not to oversell the strength of CogAct models (indeed, the models had 80% miss rate on the "Identical" metric in human simulations). In this latter sense, the current study is not a prescription of a concept learning theory, but a direction such theory should follow.

We argued that the choice of music minimised the effect of oversimplification inherent in the models that are set in linguistic domains. However, an extension of the current study may apply our methodology to the literature domain to evaluate large language models that capture meaning (see also Bubeck et al., 2023). Like with the current music study, the equivalent question of "Which works by Dickens do you recollect reading?" is similarly tractable and allows for relatively precise account of human participants' training data. The said data – in the shape of digital books – is readily available and easy to use in psychological simulations. The combination of these factors enhances validity while simultaneously simplifying both the training and the evaluation of psychological models. Indeed, the proposed extension to the current study may form an important next step in the evolution of psychological models of concept learning.

## Future Directions and Conclusions

In this article, we have focused on the perceptual aspect of concepts. We have excluded processes of concept formation that rely on verbal reasoning and logic. Furthermore, the proposed models were not trained to answer questions or to summarise information. One long-term goal would be to link the proposed general model of concept formation to natural language. This would reinforce both their continuous/fuzzy and logical/discrete nature and allow for a way to verify the grounding of semantical representations. One possibility is to integrate the proposed CogAct model of categorisation with another chunking theory based model – MOSAIC (Model Of Syntax Acquisition In Children) (Freudenthal et al., 2007; Freudenthal et al., 2009; Gegov et al., 2012). Indeed, recent advances in language models address an important potential criticism of the current study – that CogAct merely captured statistical frequency effects of the words/notes/chess pieces and not their actual meaning, a common criticism of latent semantic analysis models that confuse "dog bites man" with "man bites dog" (Braun et al., 2017). Current state-of-the-art large language models demonstrate that the relationship between meaning and frequency effects may in fact be much closer than was previously thought (Teubner et al., 2023).

More broadly, the current article served as an advertisement for the UTC approach to psychology. As was noted by Newell at the beginning of his seminal work: "psychology has arrived at the possibility of unified theories of cognition—theories that gain the power of positing a single system of mechanisms that operate together to produce the full range of human cognition. I do not say that they are here. But they are within reach and we should strive to attain them" (Newell, 1990, p. 1). The rapid advance towards general artificial intelligence not only shows the way, but also directly overlaps with psychological UTC candidates.

A common general criticism of the computational modelling approach is the potential for "overfitting" – optimising free parameters to achieve the best fit may lead to poor generalisability beyond the currently simulated data (Tetko, Livingstone, & Luik, 1995). Following Simon's (1992) advice, this study attempted to address the issue by quadrupling the ratio of data explained to free parameters used – the models used the same free parameters for the artificial category learning, literature, chess and music (including human simulations). Indeed, this approach forms the main attraction of the UTC approach which separates it from the infinite number of task-specific theories. However, conclusive guarantee that each of the models provided a unique explanation is impossible (Lakatos, 1970). An obvious expansion of CogAct is to model yet more different tasks and enrich the models to improve scores on the human-model comparison metrics.

There is an interesting intersection between the literature on concept formation and the literature on expertise; in many domains, such as chess, physics, and medicine, becoming an expert requires learning the conceptual structure of a domain, which obviously relates to its statistical structure. Being data driven, CogAct is a suitable candidate to models experiment in these domains. Among other things, this could shed light on how domain-specific knowledge facilitates the acquisition of concepts. Another potentially fruitful line of research is to investigate in detail, across a number of different tasks, the links between chunking and deep learning.

Chunking theory provides a general insight into the psychology of concept formation beyond the domains modelled in this paper. Utilising the rigour of a formal model, the CogAct architecture connects fundamental psychological structures such as attention, chunk, LTM and STM to the detailed ground-up process of adapting learning to category structure – a framework that can potentially be applied to *any* domain. Furthermore, the current study also shows CogAct's power to operationalise the factors underpinning subjectivity – learning a concept is a function of many potentially unique variables. The variation in prior knowledge, the choice and the amount of data devoted to learning a concept, and the agent's internal mechanisms – are all part of CogAct's respective computational methodologies and all play a part in predicting how individuals come to share subjective states.

We may thus conclude by paraphrasing Herbert Simon (1992): it is a justified conclusion that human concept learning and subjectivity can be characterised by operations in STM and LTM with a CogAct-like architecture, although the detailed structure of the models is open to further enrichment.